\pdfoutput=1 
\documentclass[runningheads]{llncs}
\usepackage{subcaption}
\usepackage{graphicx}
\usepackage{amsmath,amssymb} 
\usepackage{color}
\usepackage[width=122mm,left=12mm,paperwidth=146mm,height=193mm,top=12mm,paperheight=217mm]{geometry}
\usepackage{mynotation}
\usepackage{epsfig}
\usepackage{epstopdf}
\usepackage{booktabs}

\begin{document}
\pagestyle{headings}
\mainmatter

\title{Unveiling the Power of Deep Tracking}
\titlerunning{Unveiling the Power of Deep Tracking}
\authorrunning{Bhat et al.}

\author{Goutam Bhat, Joakim Johnander, Martin Danelljan, Fahad Shahbaz Khan, Michael Felsberg}
\institute{CVL, Department of Electrical Engineering, Link\"oping University, Sweden\\
\email{\{goutam.bhat, joakim.johnander, martin.danelljan, fahad.khan, michael.felsberg\}@liu.se}}

\maketitle

\begin{abstract}
	In the field of generic object tracking numerous attempts have been made to exploit deep features.
Despite all expectations, deep trackers are yet to reach an outstanding level of performance compared to methods solely based on handcrafted features.
In this paper, we investigate this key issue and propose an approach to unlock the true potential of deep features for tracking.
We systematically study the characteristics of both deep and shallow features, and their relation to tracking accuracy and robustness. We identify the limited data and low spatial resolution as the main challenges, and propose strategies to counter these issues when integrating deep features for tracking. Furthermore, we propose a novel adaptive fusion approach that leverages the complementary properties of deep and shallow features to improve both robustness \emph{and} accuracy. Extensive experiments are performed on four challenging datasets. On VOT2017, our approach significantly outperforms the top performing tracker from the challenge with a relative gain of $17\%$ in EAO.

\end{abstract}

\section{Introduction}

Generic object tracking is the problem of estimating the trajectory of a target in a video, given only its initial state. The problem is particularly difficult, primarily due to the limited training data available to learn an appearance model of the target \emph{online}. Existing methods rely on rich feature representations to address this fundamental challenge. While handcrafted features have long been employed for this task, recent focus has been shifted towards deep features. The advantages of deep features being their ability to encode high-level information, invariant to complex appearance changes and clutter.

\begin{figure}[!t]
	\centering
	\includegraphics[width=0.6\textwidth,trim={0.0cm 0.0cm 0.0cm 0.0cm},clip]{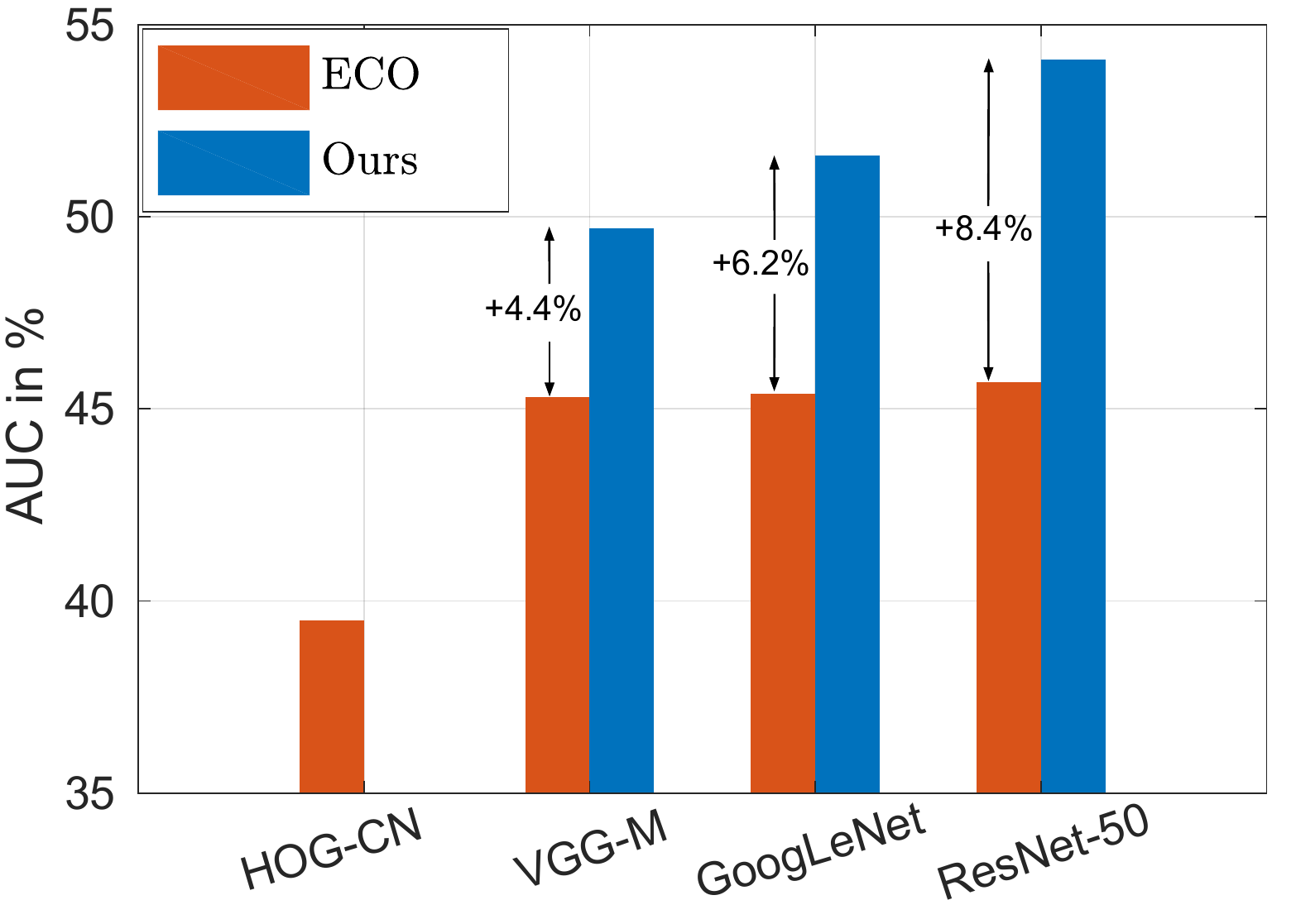}\vspace{-4mm}	
	\caption{Tracking performance on the Need for Speed dataset \cite{NfS} when using deep features extracted from different networks. In all cases, we employ the same shallow representation, consisting of HOG and Color Names. The baseline ECO \cite{DanelljanCVPR2017} does not benefit from more powerful network architectures, e.g.\ ResNet. Instead, our approach is able to exploit more powerful representations, achieving a consistent gain going from handcrafted features towards more powerful network architectures.}\vspace{-3.5mm}
	\label{fig:intro}
\end{figure}

Despite the outstanding success of deep learning in a variety of computer vision tasks, its impact in generic object tracking has been limited. In fact, trackers based on handcrafted features \cite{DanelljanCVPR2017,CSRDCF,Staple,MEEM2014} still provide competitive results, even outperforming many deep trackers on standard benchmarks \cite{OTB2015,VOT2017}. Moreover, contrary to the trend in image classification, object trackers do not tend to benefit from deeper and more sophisticated network architectures (see figure~\ref{fig:intro}). 
In this work, we investigate the reasons for the limited success of deep networks in visual object tracking.

We distinguish two key challenges generally encountered when integrating deep features into visual tracking models. Firstly, compared to traditional handcrafted approaches, it is well known that deep models are data-hungry. This becomes a major obstacle in the visual tracking scenario, where training data is extremely scarce and a robust model must be learned from a single labeled frame. Even though pre-trained deep networks are frequently employed, the target model must learn the discriminative activations possessing invariance towards \emph{unseen} appearance changes.

The second challenge for deep features is accurate target prediction. Not only is precise target localization crucial for tracking performance, it also affects the learning of the model since new frames are annotated by the tracker itself. As a result, inaccurate predictions may lead to model drift and eventual tracking failure. Deep convolutional layers generally trade spatial resolution for increased high-level invariance to account for appearance changes. Consequently, many trackers complement the deep representation with shallow-level activations \cite{DanelljanECCV2016,HCF_ICCV15} or handcrafted features \cite{DanelljanCVPR2017} for improved localization accuracy. This raises the question of how to optimally fuse the fundamentally different properties of shallow and deep features in order to achieve both accuracy \emph{and} robustness.

\noindent\textbf{Contributions:} 
In this paper, we analyze the influential characteristics of deep and shallow features for visual tracking. This is performed by (i) systematically studying the impact of a variety of data augmentation techniques and (ii) investigating the accuracy-robustness trade-off in the discriminative learning of the target model. Our findings suggest that extensive data augmentation leads to a remarkable performance boost for the deep-feature-based model while negatively affecting its shallow counterpart. Furthermore, we find that the deep model should be trained for robustness, while the shallow model should emphasize accurate target localization. These results indicate that the deep and shallow models should be trained \emph{independently} and fused at a later stage.
As our second contribution, we propose a novel fusion strategy to combine the deep and shallow predictions in order to exploit their complementary characteristics. This is obtained by introducing a quality measure for the predicted state, taking both accuracy and robustness into account.

Experiments are performed on five challenging benchmarks: Need for Speed, VOT2017, Temple128, UAV123 and OTB-2015. 
Our results clearly demonstrate that the proposed approach provides a significant improvement over the baseline tracker. Further, our approach sets a new state-of-the-art on all four tracking datasets. On the VOT2017 benchmark, our approach achieves an EAO score of $0.378$, surpassing the competition winners ($0.323$) with a relative gain of $17\%$.

\section{Related Work}

Deep learning has pervaded many areas of computer vision. While these techniques have also been investigated for visual tracking, it has been with limited success.
The SINT method~\cite{Tao2016Sint} learns a similarity measure on offline videos, and localize the target using the initial labeled sample. Li et al.~\cite{LiBMVC14} tackle the tracking problem in an end-to-end fashion by training a classifier online. The FCNT~\cite{Lijun15ICCV} employs both pre-trained deep features and an online trained model. MDNet~\cite{MDNet} further pre-trains a model offline using a multi-domain procedure. The offline pre-training requirement was later removed in TCNN~\cite{TCNN}. Following the end-to-end philosophy, recent works~\cite{Valmadre2017cvpr,Song2017CREST} have investigated integrating Discriminative Correlation Filters (DCF) \cite{MOSSE2010,DanelljanICCV2015} as a computational block in a deep network. The work of \cite{Valmadre2017cvpr} integrate DCF into the Siamese framework~\cite{SiameseFC}. Further, \cite{Song2017CREST} employs DCF as a one-layer CNN for end-to-end training.

Other DCF methods focus on integrating convolutional features from a fixed pre-trained deep network.
Ma et al.\ \cite{HCF_ICCV15} propose a hierarchical ensemble method of independent DCF trackers to combine multiple convolutional layers.
Qi et al.~\cite{Qi2016Hedge} learn a correlation filter per feature map, and combine the individual predictions with a modified Hedge algorithm. The MCPF tracker proposed by Zhang et al.~\cite{ZhangCVPR17MCPF} combines the deep DCF with a particle filter.
Danelljan et al.\ ~\cite{DanelljanECCV2016} propose the continuous convolution operator tracker (C-COT) to efficiently integrate multi-resolution shallow and deep feature maps.  
The subsequent ECO tracker \cite{DanelljanCVPR2017} improves the C-COT tracker in terms of performance and efficiency. In this work we adopt the ECO tracking framework due to its versatility and popularity: in the most recent edition of VOT2017 \cite{VOT2017}, five of the top 10 trackers were based on either ECO or its predecessor C-COT.

\section{Analyzing Deep Features for Tracking} 
\label{sec:analysis}

Deep learning has brought remarkable performance improvements in many computer vision areas, such as object classification, detection and semantic segmentation. However, its impact is yet to be \emph{fully} realized in the context of generic visual object tracking. In this section, we analyze the causes behind the below-expected performance of deep trackers and propose strategies to address them.

\subsection{Motivation}

In our quest to seek a better understanding about deep features for tracking, we investigate their properties in relation to the well studied shallow features. One of the well known issues of deep learning is the need for large amounts of labeled training data. Still, thousands of training samples are required to \emph{fine-tune} a pre-trained deep network for a new task. Such amount of data is however not available in the visual tracking scenario, where initially only a single labeled frame is provided. This poses a major challenge when learning deep-feature-based models for visual tracking.

To maximally exploit the available training data, deep learning methods generally employ data augmentation strategies. Yet, data augmentation is seldom used in visual tracking. In fact, the pioneering work of Bolme et al.~\cite{MOSSE2010} utilized augmented gray-scale image samples to train a discriminative tracking model. Since then, state-of-the-art deep DCF tracking methods have ignored data augmentation as a strategy for acquiring additional training data. 
 In section~\ref{sec:data_augmentation} we therefore perform a thorough investigation of data augmentation techniques with the aim of better understanding deep features for tracking.

Another challenge when integrating deep features is their low spatial resolution, hampering accurate localization of the target. Object trackers based on low-level handcrafted features are primarily trained for accurate target localization to avoid long-term drift. However, this might not be the optimal strategy for deep features which exhibit fundamentally different properties. Deep features generally capture high-level semantics while being invariant to, e.g., small translations and scale changes. From this perspective it may be beneficial to train the deep model to emphasize robustness rather than accuracy. This motivates us to analyze the accuracy/robustness trade-off involved in the model learning, to gain more knowledge about the properties of deep and shallow features. This analysis is performed in section~\ref{sec:robustness_vs_accuracy}.

\subsection{Methodology}

To obtain a clearer understanding of deep and shallow features, we aim to isolate their impact on the overall tracking performance. The analysis is therefore performed with a baseline tracker that exclusively employs either shallow or deep features. This exclusive treatment allows us to directly measure the impact of, e.g., data augmentation on both shallow and deep features separately.

We use the recently introduced ECO tracker \cite{DanelljanCVPR2017} as a baseline, due to its state-of-the-art performance. For shallow features, we employ a combination of Histogram of Oriented Gradients (HOG) \cite{Dalal05} and Color Names (CN) \cite{Weijer09a}, as it has been used in numerous tracking approaches \cite{DanelljanCVPR2017,DanelljanICCV2015,li2014scale,hong2015multi,lukevzivc2016discriminative}. For the deep representation, we first restrict our analysis to ResNet-50, using the activations from the fourth convolutional block. Generalization to other networks is further presented in section~\ref{sec:other_networks}. The entire analysis is performed on the OTB-2015~\cite{OTB2015} dataset.

\subsection{Data Augmentation}
\label{sec:data_augmentation}

Data augmentation is a standard strategy to alleviate problems with limited training data. It can lead to a better generalization of the learned model for unseen data. However, data augmentation can also lead to lower accuracy in the context of visual tracking due to increased invariance of the model to small translations or scale changes. Therefore, it is unclear whether data augmentation is helpful in the case of tracking.

We separately investigate the impact of different data augmentation techniques on both shallow as well as deep features. We consider the following data augmentation techniques:\\
\noindent\textbf{Flip:} The sample is horizontally flipped.\\
\noindent\textbf{Rotation:} Rotation from a fixed set of $12$ angles ranging from $-60^\circ$ to $60^\circ$.\\ 
\noindent\textbf{Shift:} Shift of $n$ pixels both horizontally and vertically prior to feature extraction. The resulting feature map is shifted back $n/s$ pixels where $s$ is the stride of the feature extraction.\\
\noindent\textbf{Blur:} Blur with a Gaussian filter. This is expected to simulate motion blur and scale variations, which are both commonly encountered in tracking scenarios.\\
\noindent\textbf{Dropout:} Channel-wise dropout of the sample. This is performed by randomly setting $20\%$ of the feature channels to zero. As usual, the remaining feature channels are amplified in order to preserve the sample energy. 

\begin{figure}[t]
	\begin{subfigure}[t]{0.5\textwidth}
		\includegraphics*[width=1\textwidth,trim={0.0cm 0.0cm 1.0cm 0.0cm},clip]{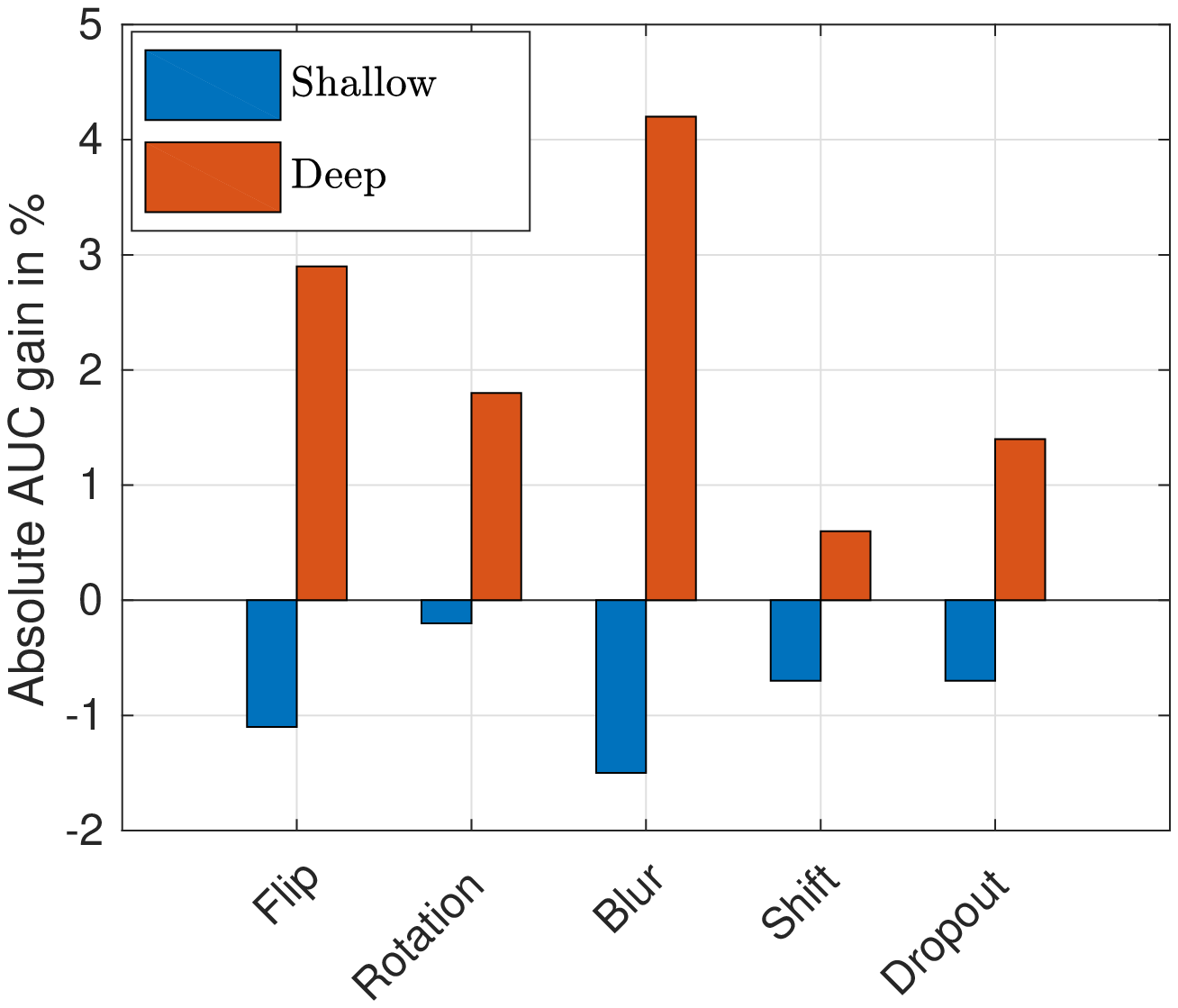}
		\caption{Impact of data augmentation}
		\label{fig:data_aug_expt}
	\end{subfigure}%
	\begin{subfigure}[t]{0.5\textwidth}
		\includegraphics*[width=1\textwidth,trim={0.0cm 0.0cm 1.0cm 0.0cm},clip]{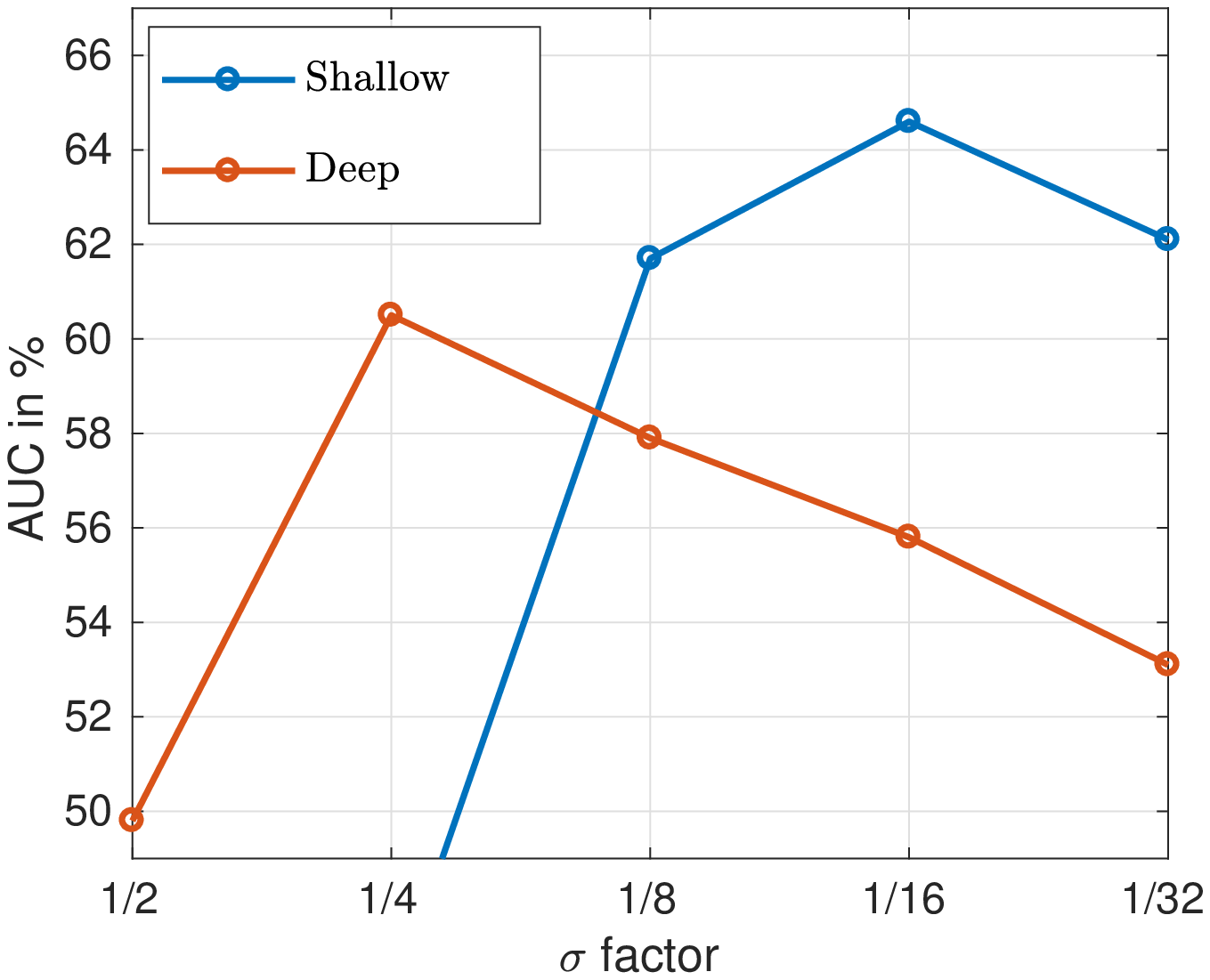}
		\caption{Impact of label function}
		\label{fig:sigma_expt}
	\end{subfigure}\vspace{-2mm}
	\caption{Impact of data augmentation (a) and label function (b) on shallow (blue) and deep (red) features on the OTB-2015 dataset. The results are reported as area-under-the-curve (AUC). While deep features significantly benefit from data augmentation, the results deteriorate for shallow features. Similarly, a sharp label function is beneficial for the shallow features, whereas the deep features benefit from a wide label function.}\vspace{-3mm}
	\label{fig:analysis}
\end{figure}

Figure \ref{fig:data_aug_expt} shows the impact of data augmentation on tracking performance (in AUC score \cite{OTB2015}). It can be seen that the deep features consistently benefit from data augmentation. All augmentations, except for 'shift', give over $1\%$ improvement in tracking performance. The maximum improvement is obtained using 'blur' augmentation, where a gain of $4\%$ is obtained over the baseline, which employs no data augmentation. Meanwhile, shallow features are adversely affected by data augmentation, with all augmentation types leading to lower tracking performance. This surprising difference in behavior of deep and shallow features can be explained by their opposing properties. Deep features capture higher level semantic information that is invariant to the applied augmentations like 'flip', and can thus gain from the increased training data. On the other hand, the shallow features capture low-level information that is hampered by augmentations like 'flip' or 'blur'. The use of data augmentation thus harms the training in this case.

\subsection{Robustness/Accuracy Trade-off}
\label{sec:robustness_vs_accuracy}
When comparing the performance of trackers, there are two important criteria: accuracy and robustness. The former is the measure of \emph{how accurately} the target is localized during tracking. Robustness, on the other hand, is the tracker's resilience to failures in challenging scenarios and its ability to recover. In other words, robustness is a measure of \emph{how often} the target is successfully localized. Generally, both accuracy and robustness are of importance, and a satisfactory trade-off between these properties is sought since they are weakly correlated \cite{KristanPAMI2016}. This trade-off can be controlled in the construction and training of the tracker.

In a discriminative tracking framework, the appearance model can be learned to emphasize the accuracy criterion by only extracting positive samples very close to the target location. That is, only very accurate locations are treated as positive samples of the target appearance. Instead, increasing the region from which target samples are extracted allows for more positive training data. This has the potential of promoting the generalization and robustness of the model, but can also result in poor discriminative power when the variations in the target samples become too large.

We analyze the effect of training the tracking model for various degrees of accuracy-robustness trade-off when using either shallow or deep features. In DCF-based trackers, such as the baseline ECO, the size of the region from which positive training samples are extracted is controlled by the width of the label detection score. ECO employs a Gaussian function for this task, with standard deviation proportional to the target size with a factor of $\sigma$. We analyze different values of $\sigma$ for both shallow and deep features.
Figure \ref{fig:sigma_expt} shows the results of this experiment. We observe that the deep features are utilized best when trained with higher value of $\sigma$, with $\sigma=\frac{1}{4}$ giving the best results. This behavior can be attributed to the invariance property of the deep features. Since they are invariant to small translations, training deep features to get higher accuracy might lead to a suboptimal model. The shallow features on the other hand, perform best when trained with a low $\sigma$ and give inferior results when trained with a higher $\sigma$. This is due to the fact that the shallow features capture low level, higher resolution features, and hence are well suited to give high accuracy. Furthermore, due to their large variance to small transformations, the model is unable to handle the larger number of positive training samples implied by a higher $\sigma$, resulting in poor performance.

\begin{table}[!t]
	\centering
        \caption{Impact of data augmentation (denoted Aug) and wider label score function (denoted $\sigma$) on deep features. Results are shown in terms of AUC score on the OTB-2015 dataset. Both data augmentation and wider label scores provide significant improvement. The best result is obtained when employing both techniques.}
	\resizebox{0.65\columnwidth}{!}{%
		\begin{tabular}{l@{~}c@{~~}c@{~~}c@{~~}c@{~~}c@{~~}}
\toprule
&ResNet&ResNet+Aug&ResNet+$\sigma$&ResNet+Aug+$\sigma$\\\midrule
AUC&56.2&61.5&60.5&\textbf{62.0}\\\bottomrule
\end{tabular}

	}\vspace{1mm}%
	\label{tab:deep_baseline}%
	\vspace{-3mm}
\end{table}

\subsection{Observations}
\label{sec:observations}
The aforementioned results from section~\ref{sec:data_augmentation} and section \ref{sec:robustness_vs_accuracy} show that the deep model significantly improves by the use of data augmentation \emph{and} by training for increased robustness instead of accuracy. We further evaluate the combined effects of data augmentation and higher $\sigma$ on the deep model. Table~\ref{tab:deep_baseline} shows the results on the OTB-2015 dataset in terms of the AUC measure. The baseline tracker (left) does not employ data augmentation and uses the default value $\sigma = \frac{1}{12}$. Combining all the data augmentation techniques evaluated in section~\ref{sec:data_augmentation} provides an improvement of $5.3\%$ in AUC over the baseline. Training with a $\sigma$-parameter of $\frac{1}{4}$ further improves the results by $0.5\%$. 
Thus our analysis indicates the benefit of using both data augmentation as well as a wider label function when training the deep-feature-based model. 

Results from section \ref{sec:data_augmentation} and section \ref{sec:robustness_vs_accuracy} thus highlight the complementary properties of deep and shallow features. Their corresponding models need to be trained differently, in terms of data and annotations, in order to best exploit their true potential. We therefore argue that the shallow and deep models should be trained independently. However, this raises the question of how to fuse these models in order to leverage their complementary properties, which we address in the next section.

\section{Adaptive Fusion of Model Predictions}
\label{sec:fusion}

\begin{figure}[t]
	\centering
	\subcaptionbox{Image sample}{\includegraphics[width=0.25\textwidth,trim={1.5cm 4cm 7.05cm 1cm},clip]{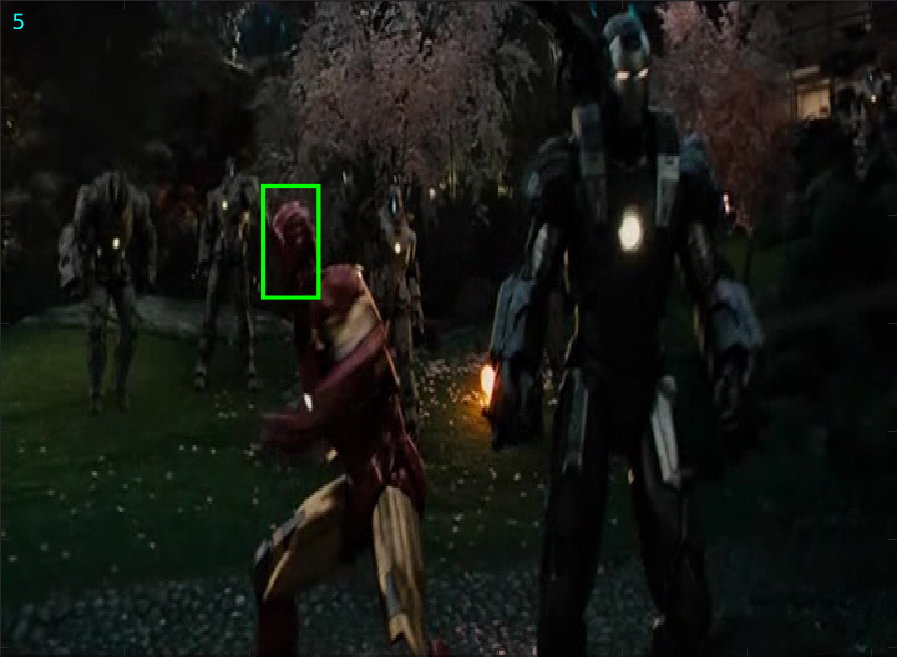}}%
	\subcaptionbox{Deep score}{\includegraphics*[width=0.25\textwidth,trim={2.8cm 1.3cm 2.8cm 1.3cm},clip]{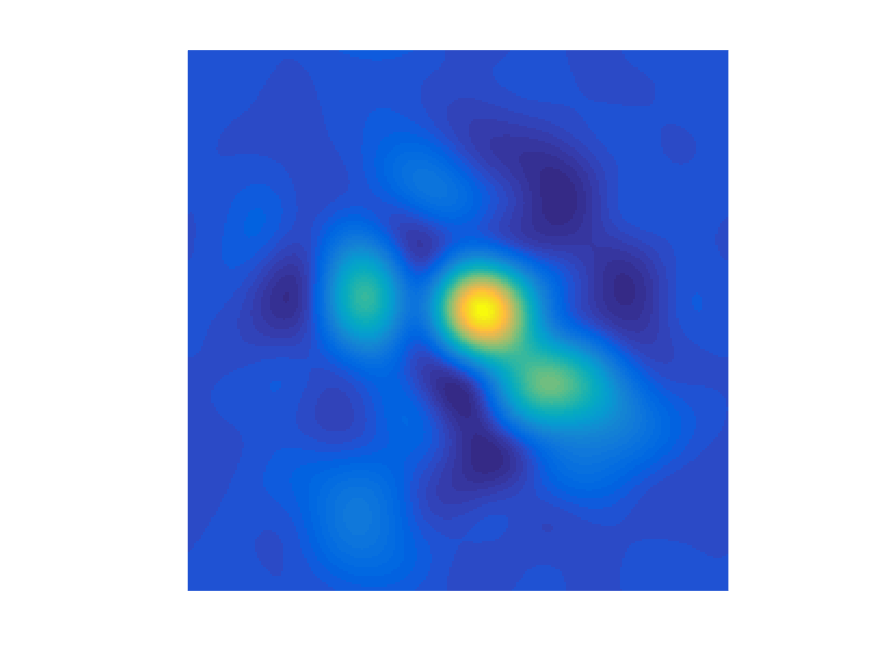}}%
	\subcaptionbox{Shallow score}{\includegraphics[width=0.25\textwidth,trim={2.8cm 1.3cm 2.8cm 1.3cm},clip]{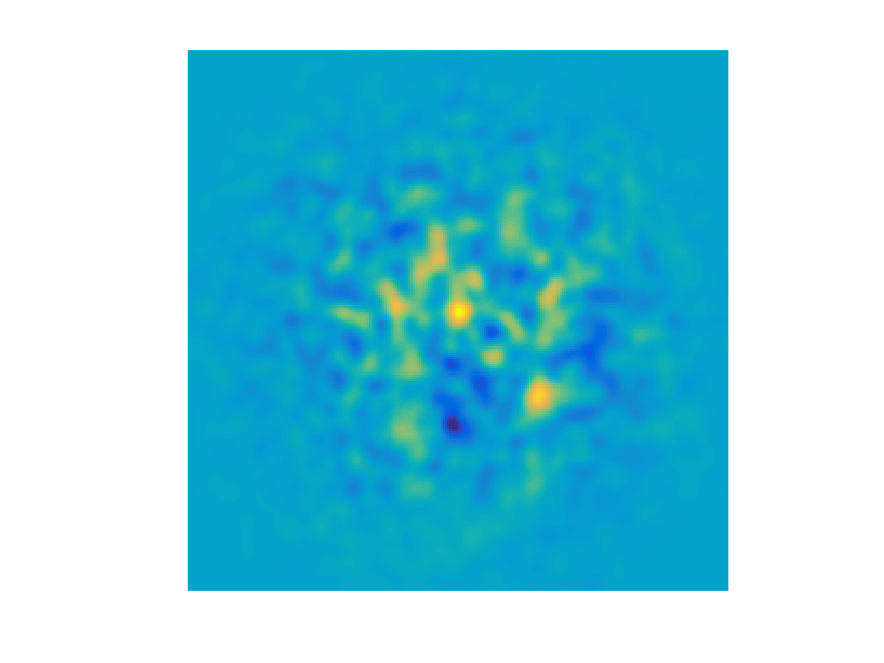}}%
	\subcaptionbox{Fused score}{\includegraphics[width=0.25\textwidth,trim={2.8cm 1.3cm 2.8cm 1.3cm},clip]{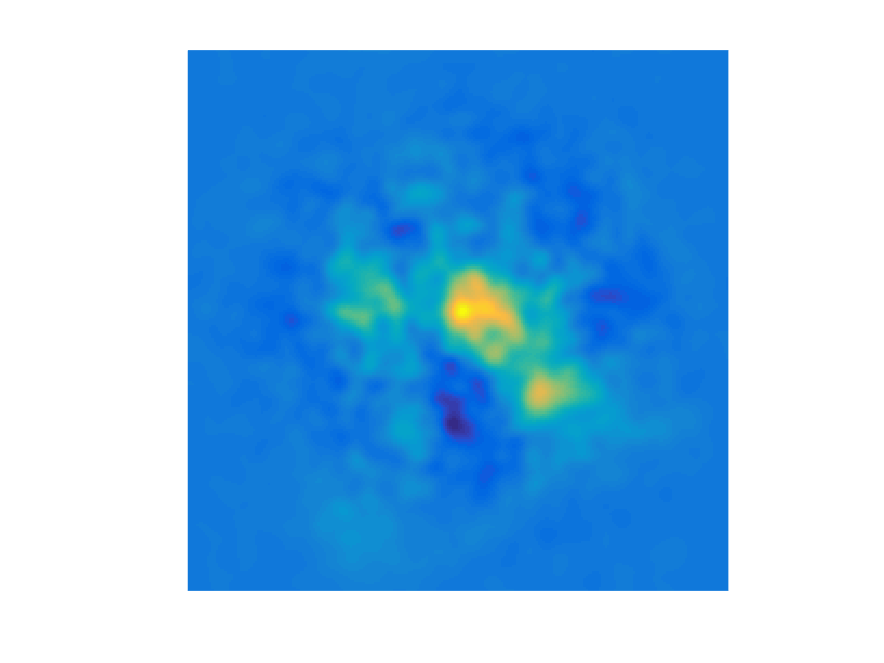}}%
	\vspace{-2mm}
	\caption{Visualization of the detection scores produced by the deep and shallow models for a sample frame (a). The deep score (b) contains a robust mode with high confidence, but it only allows for coarse localization. Meanwhile, the shallow score (c) have sharp peaks enabling accurate localization, but does also contains distractor modes. Our approach fuses these scores by adaptively finding the optimal weights for each model, producing a sharp and unambiguous score function (d).}\vspace{-3mm}
	\label{fig:scores}
\end{figure}

As previously discussed, the deep and shallow models possess different characteristics regarding accuracy and robustness.
This is demonstrated in figure~\ref{fig:scores}, showing detection scores from the shallow and deep models for an example frame. We propose a novel adaptive fusion approach that aims at fully exploiting their complementary nature, based on a quality measure described in section~\ref{sec:quality_measure}. In section~\ref{sec:fusion_inference} we show how to infer the deep and shallow weights and obtain the final target prediction.

\subsection{Prediction Quality Measure}
\label{sec:quality_measure}

Our aim is to find a quality measure for the target prediction, given the detection score $y$ over the search region of the image. We see the score $y$ as a function over the image coordinates, where $y(t) \in \reals$ is the target prediction score at location $t \in \reals^2$. We require that the quality measure rewards both the accuracy and robustness of the target prediction. The former is related to the \emph{sharpness} of the detection score around the prediction. A sharper peak indicates more accurate localization capability. The robustness of the prediction is derived from the margin to distractor peaks. If the margin is small, the prediction is ambiguous. On the other hand, a large margin indicates that the confidence of the prediction is significantly higher than at other candidate locations.

We propose the minimal weighted confidence margin as a quality measure of the target prediction $t^*$,
\begin{align}
  \xi_{t^*}\{y\} = \min_{t}\frac{y(t^*)-y(t)}{\Delta(t-t^*)}\enspace.
  \label{eq:xi}
\end{align}
The confidence margin in the numerator is weighted by a distance measure $\Delta : \reals^2 \rightarrow [0,1]$ satisfying $\Delta(0) = 0$ and \mbox{$\lim_{|\tau| \rightarrow \infty} \Delta(\tau) = 1$}. We also assume $\Delta$ to be twice continuously differentiable and have a positive definite Hessian at $\tau = 0$. For our purpose, we use the function,
\begin{align}
  \Delta(\tau) = 1 - e^{-\frac{\kappa}{2}|\tau|^2}\enspace.
  \label{eq:delta_func}
\end{align}
Here, $\kappa$ is a parameter controlling the rate of transition $\Delta(\tau) \rightarrow 1$ when $|\tau|$ is increasing. As we will see, $\kappa$ has a direct interpretation related to the behavior of the quality measure \eqref{eq:xi} close to the target prediction $t \approx t^*$. From the definition \eqref{eq:xi} it follows that $\xi_{t^*}\{y\} \geq 0$ if and only if $y(t^*)$ is a global maximum of $y$.

To verify that the proposed quality measure \eqref{eq:xi} has the desired properties of promoting both accuracy and robustness, we analyze the cases (a) where $t$ is far from the prediction $t^*$ and (b) when $t \rightarrow t^*$. In the former case, we obtain $|t - t^*| \gg 0$ implying that $\Delta(t - t^*) \approx 1$. In this case, we have that
\begin{equation}
	\label{eq:far-bound}
	\xi_{t^*}\{y\}  \leq \frac{y(t^*)-y(t)}{\Delta(t-t^*)} \approx y(t^*)-y(t) \,, \quad \text{whenever} \,\, |t - t^*| \gg 0 \enspace.
\end{equation}
As a result, the quality measure $\xi_{t^*}\{y\}$ is approximately bounded by the score-difference to the most significant distractor peak $y(t)$ outside the immediate neighborhood of the prediction $t^*$. Hence, a large quality measure $\xi_{t^*}\{y\}$ ensures that there are no distractors making the prediction ambiguous. Conversely, if there exists a secondary detection peak $y(t)$ with a similar score $y(t) \approx y(t^*)$, then the quality of the prediction is low $\xi_{t^*}\{y\} \approx 0$. 

In the other case we study how the measure \eqref{eq:xi} promotes an accurate prediction by analyzing the limit $t \rightarrow t^*$. We assume that the detection score function $y$ is defined over a continuous domain $\Omega \subset \reals^2$ and is twice continuously differentiable. This assumption is still valid for discrete scores $y$ by applying suitable interpolation. The ECO framework in fact outputs scores with a direct continuous interpretation, parametrized by its Fourier coefficients. In any case, we assume the prediction $t^*$ to be a local maximum of $y$. We denote the gradient and Hessian of $y$ at $t$ as $\nabla y(t)$ and $\hess y(t)$ respectively. Since $t^*$ is a local maximum we conclude $\nabla y(t^*) = 0$ and $0 \geq \lambda_1^* \geq \lambda_2^*$, where $\lambda_1^*, \lambda_2^*$ are the eigenvalues of $\hess y(t^*)$. Using \eqref{eq:delta_func}, we obtain the result\footnote{See the supplementary material for a derivation.}
\begin{align}
	\xi_{t^*}\{y\} \le \frac{|\lambda_1^*|}{\kappa} \,.
	\label{eq:close-bound}
\end{align}

Note that the eigenvalue $|\lambda_1^*|$ represents the minimum curvature of the score function $y$ at the peak $t^*$. Thus, $|\lambda_1^*|$ is a measure of the \emph{sharpness} of the peak $t^*$. The quality bound \eqref{eq:close-bound} is proportional to the sharpness of the scores at the prediction. A high quality value $\xi_{t^*}\{y\}$ ensures that the peak is distinctive, while a flat peak will result on a low quality value. The parameter $\kappa$ controls the trade-off between the promotion of robustness and accuracy of the prediction. From \eqref{eq:close-bound} it follows that $\kappa$ represents the sharpness $|\lambda_1^*|$ that yields a quality of at most $\xi_{t^*}\{y\} = 1$.

Our approach can be generalized to scale transformations and other higher-dimensional state spaces by extending $t$ to the entire state vector. In this paper, we employ 2-dimensional translation and 1-dimensional scale transformations. In the next section, we show that \eqref{eq:xi} can be used for jointly finding the prediction $t^*$ and the optimal importance weights for the shallow and deep scores.

\begin{figure}[t]
	\centering
	\includegraphics[width=0.45\textwidth,trim={0cm 0cm 0cm 0cm},clip]{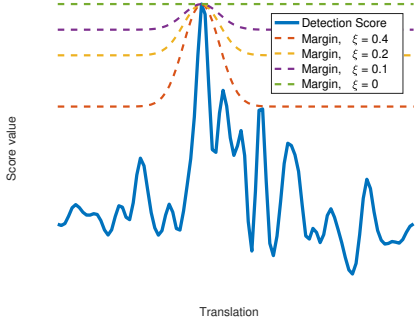}\vspace{-2mm}
	\caption{An illustration of our fusion approach, based on solving the optimization problem \eqref{eq:fusionopt2}. A one-dimensional detection score $y_\beta(t)$ (blue curve) is plotted for a particular choice of model weights $\beta$, with the candidate state $t^*$ corresponding to the global maximum. The left-hand side of \eqref{eq:slackconstraint} (dashed curves) is plotted for different values of the slack variable $\xi$, representing the margin. We find the maximum value of $\xi$ satisfying the inequality \eqref{eq:slackconstraint}, which in this case is $\xi = 0.4$.}\vspace{-3mm}
	\label{fig:distinctiveness}
\end{figure}

\subsection{Target Prediction}
\label{sec:fusion_inference}

We present a fusion approach based on the quality measure \eqref{eq:xi}, that combines the deep and shallow model predictions to find the optimal state.
Let $y_\deep$ and $y_\shallow$ denote the scores based on deep and shallow features respectively. The fused score is obtained as a weighted combination of the two scores
\begin{align}
\label{eq:fusionsum}
  y_\beta(t) = \beta_\deep y_\deep(t) + \beta_\shallow y_\shallow(t) \enspace,
\end{align}
where $\beta=(\beta_\deep,\beta_\shallow)$ are the weights for the deep and shallow scores, respectively. Our aim is to jointly estimate the score weights $\beta$ and the target state $t^*$ that maximize the quality measure \eqref{eq:xi}. This is achieved by minimizing the loss
\begin{subequations}
\label{eq:fusionopt}
  \begin{align}
    \text{minimize:} \quad\quad & L_{t^*}(\beta) = -\xi_{t^*}\{y_\beta\} + \mu\left(\beta_\deep^2 + \beta_\shallow^2\right)\\
    \text{subject to:} \quad\quad & \beta_\deep+\beta_\shallow = 1 \,,\quad \beta_\deep \geq 0 \,,\quad \beta_\shallow \geq 0\enspace.
  \end{align}
\end{subequations}
Note that we have added a regularization term, controlled by the parameter $\mu$, penalizing large deviations in the weights. The score weights themselves are constrained to be non-negative and sum up to one.

To optimize \eqref{eq:fusionopt}, we introduce a slack variable $\xi = \xi_{t^*}\{y_\beta\}$, resulting in the equivalent minimization problem
\begin{subequations}
\label{eq:fusionopt2}
  \begin{align}
    \text{minimize:} \quad\quad & L_{t^*}(\xi,\beta) = -\xi + \mu\left(\beta_\deep^2 + \beta_\shallow^2\right) \label{eq:loss2}\\
    \text{subject to:} \quad\quad & \beta_\deep+\beta_\shallow = 1 \,,\quad \beta_\deep \geq 0 \,,\quad \beta_\shallow \geq 0 \\
    & y_\beta(t^*) - \xi \Delta(t^*-t) \geq y_\beta(t) \,,\quad \forall t \in \Omega \label{eq:slackconstraint} \enspace.
  \end{align}
\end{subequations}
A visualization of this reformulated problem and the constraint \eqref{eq:slackconstraint} is shown in figure~\ref{fig:distinctiveness}. For any fixed state $t^*$, ~\eqref{eq:fusionopt2} corresponds to a Quadratic Programming (QP) problem, which can be solved using standard techniques. In practice, we sample a finite set of candidate states $\Omega$ based on local maxima from the deep and shallow scores. Subsequently, \eqref{eq:fusionopt2} is optimized for each state $t^* \in \Omega$. This adds little computational overhead, as each QP subproblem is of only three variables. We then select the candidate state $t^*$ with the lowest overall loss \eqref{eq:loss2} as our final prediction of the target.

\section{Experiments}

\subsection{Implementation Details}
Our tracker is implemented in Matlab using MatConvNet \cite{matconvnet}.
Based on the analysis in section~\ref{sec:robustness_vs_accuracy}, we select $\sigma_\deep = 1/4$ and $\sigma_\shallow = 1/16$ for deep and shallow label functions respectively, when training the models. 
As concluded in section~\ref{sec:data_augmentation}, we employ the proposed data augmentation techniques only for deep features. For the fusion method presented in section~\ref{sec:fusion}, the regularization parameter $\mu$ in \eqref{eq:fusionopt} is set to 0.15. We set the $\kappa$ parameter in the distance measure \eqref{eq:delta_func} to be inversely proportional to the target size with a factor of $8$. All parameters were set using a \emph{separate} validation set, described in the next section. We then use the same set of parameters for all datasets, throughout all experiments.

\subsection{Evaluation Methodology}

We evaluate our method on four challenging benchmarks: the recently introduced Need For Speed (NFS)~\cite{NfS}, VOT2017~\cite{VOT2017}, UAV123~\cite{UAV123}, and Temple128~\cite{TempleColor}. NFS consists of 100 videos captured using high frame rate (240 fps) cameras as well as their 30 fps versions. We use the 30 fps version of the dataset for our experiments. The dataset uses mean overlap precision (OP) and area-under-the-curve (AUC) scores to rank trackers. The OP score is computed as the percentage of frames in a video where the intersection-over-union (IOU) overlap with the ground-truth exceeds a certain threshold. The mean OP over all videos is plotted over the range of IOU thresholds $[0, 1]$ to get the success plot. The area under this plot gives the AUC score. We refer to~\cite{OTB2015} for details. Due to the stochastic nature of the dropout augmentation, we run our tracker 10 times on each sequence and report average scores to robustly estimate the performance on \emph{all} datasets. Details about VOT2017, UAV123 and Temple128 are provided in section~\ref{sec:sota}.

\noindent\textbf{Validation set:}
We use a subset of the popular OTB-2015 dataset \cite{OTB2015} as our validation set for tuning all hyperparameters. The OTB-2015 dataset has been commonly used for evaluation by the tracking community. However, the dataset has saturated in recent years with several trackers \cite{DanelljanCVPR2017,MDNet} achieving over $90\%$ OP score at threshold $0.5$ due to the majority of relatively easy videos. Instead, we are primarily interested in advancing tracking performance in the challenging and unsolved cases, where deep features are of importance. 
We therefore construct a subset of hard videos from OTB-2015 to form our validation set, termed OTB-H. 
To find the \emph{hardest} videos in OTB-2015, we consider the per-video results of four deep-feature-based trackers with top overall performance on the dataset: ECO \cite{DanelljanCVPR2017}, C-COT \cite{DanelljanECCV2016}, MDNet \cite{MDNet}, and TCNN \cite{TCNN}. We first select sequences for which the average IOU is less than $0.6$ for at least two of the four trackers. We further remove sequences overlapping with the VOT2017 dataset. The resulting OTB-H contains 23 sequences, which we use as the validation set when setting all the parameters. The remaining 73 easier videos form the OTB-E dataset that we use in our ablative studies as a test set along with NFS dataset.\footnote{See the supplementary material for a full listing of OTB-H and OTB-E.}

\subsection{Ablative Study}

\begin{figure}[t]
	\centering
	\includegraphics*[width=0.25\textwidth,trim={0.0cm 0.0cm 0.9cm 0.0cm},clip]{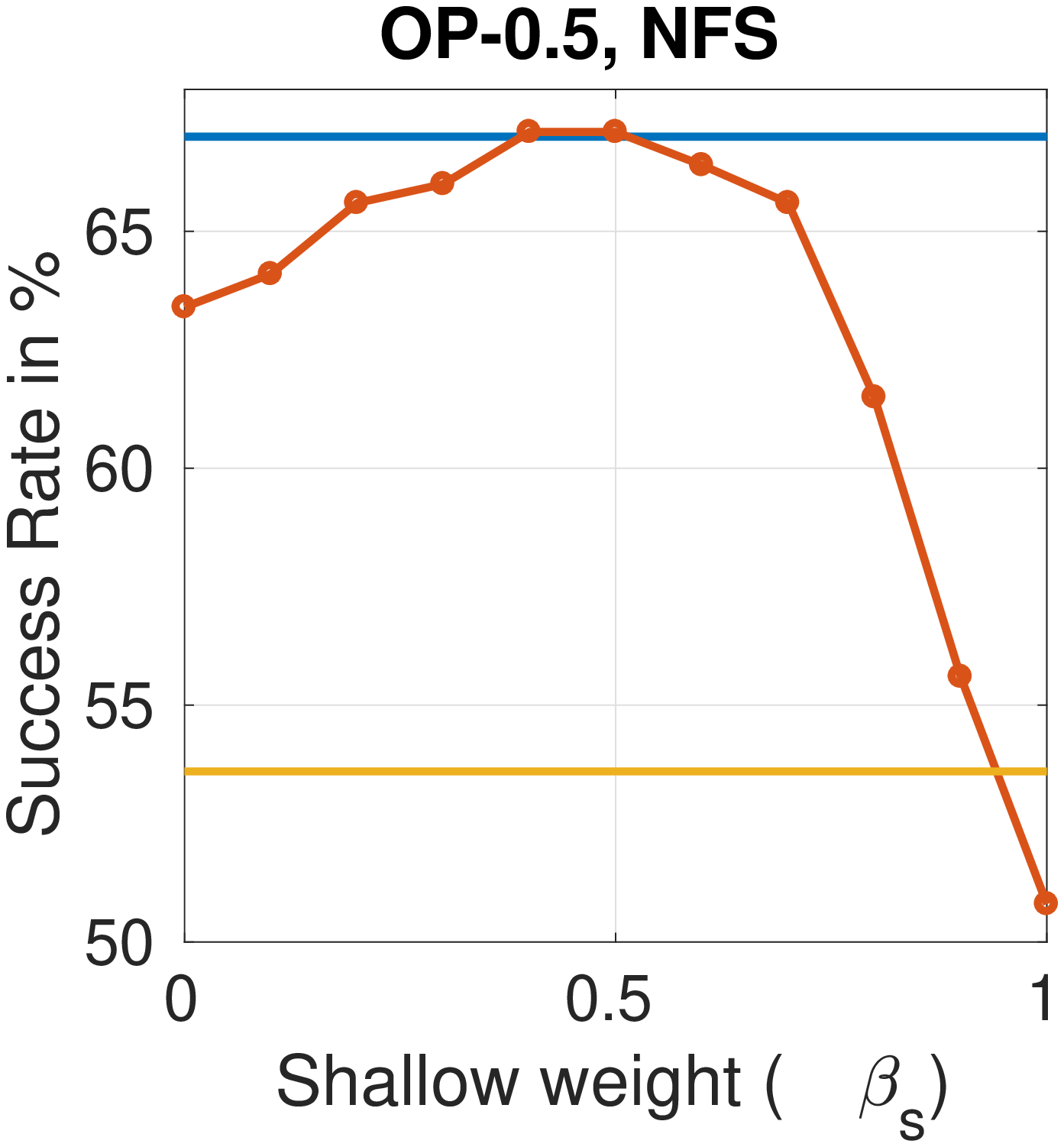}%
	\includegraphics*[width=0.25\textwidth,trim={0.0cm 0.0cm 0.9cm 0.0cm},clip]{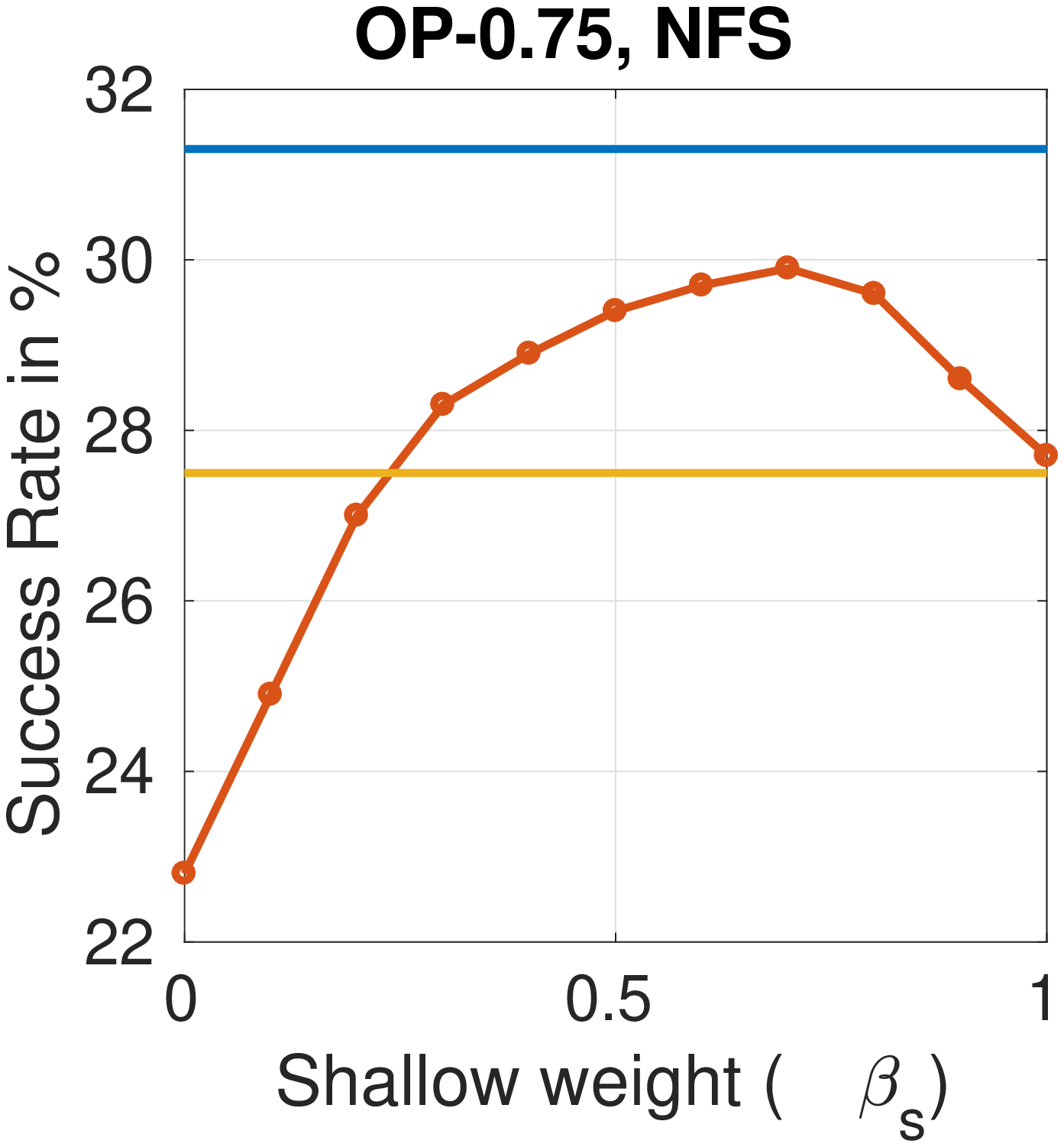}%
	\includegraphics*[width=0.25\textwidth,trim={0.0cm 0.0cm 0.9cm 0.0cm},clip]{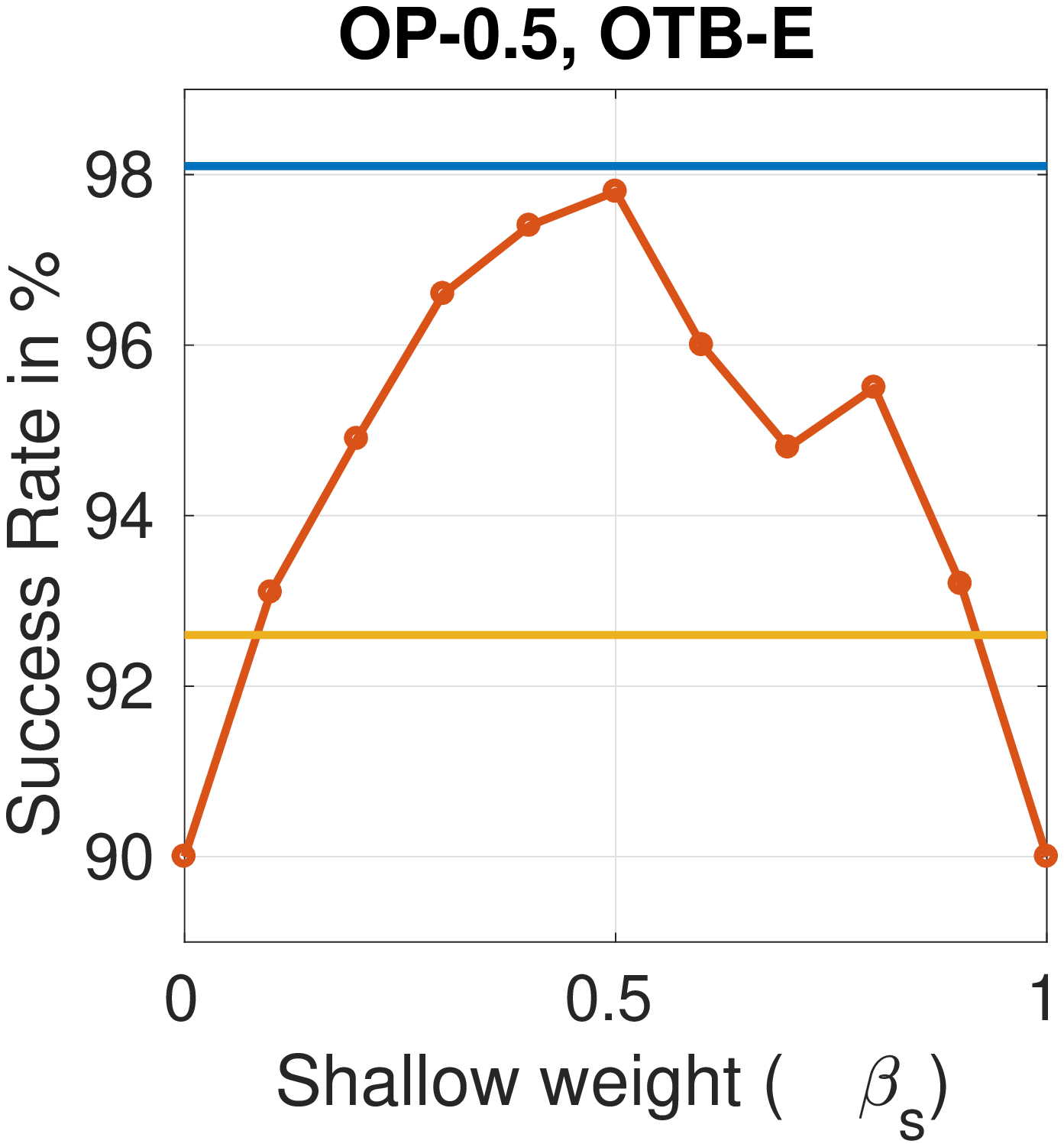}%
	\includegraphics*[width=0.25\textwidth,trim={0.0cm 0.0cm 1.0cm 0.0cm},clip]{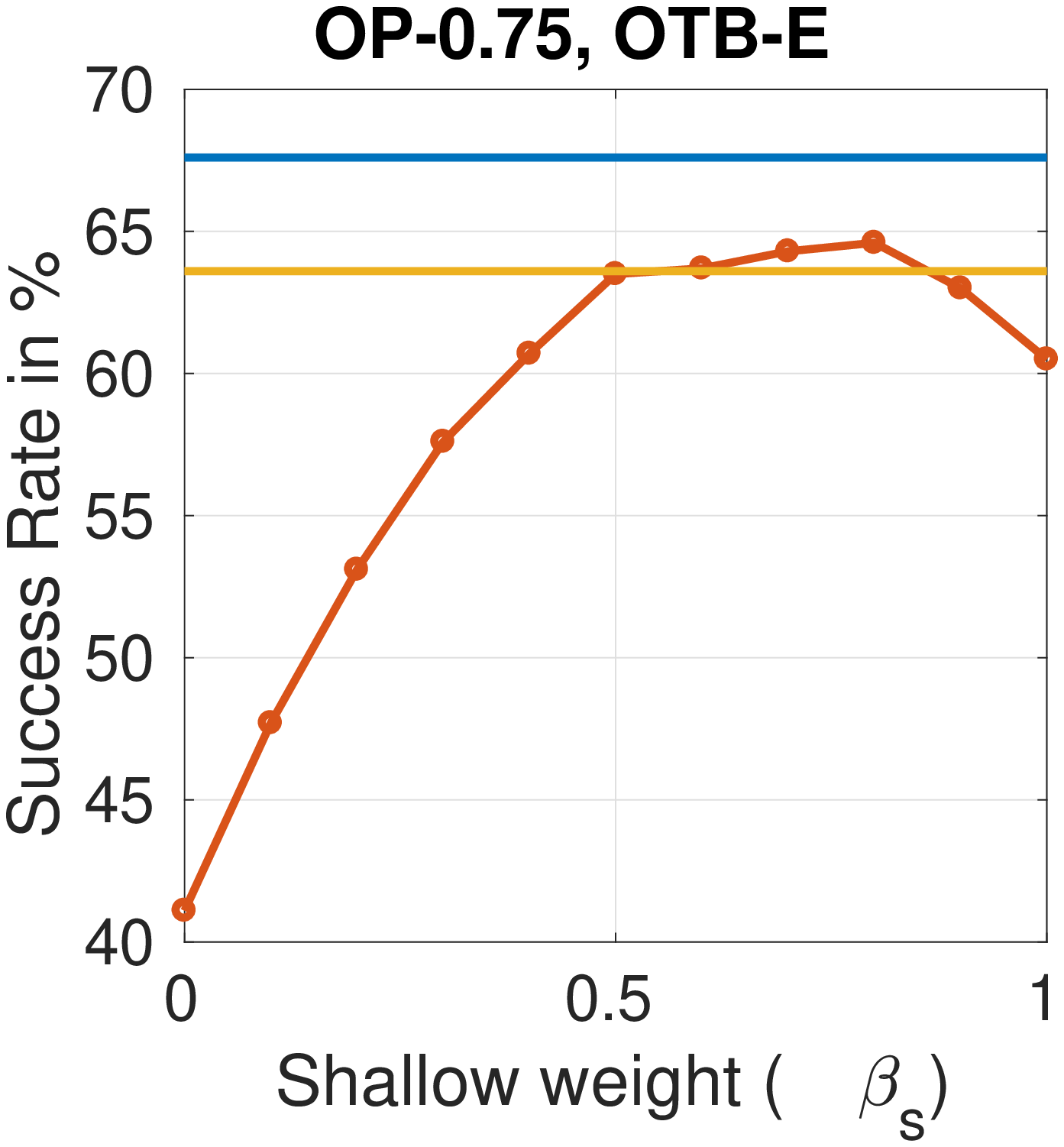}
	\includegraphics[width=0.4\textwidth,trim={0.0cm 0.0cm 0.0cm 0.0cm},clip]{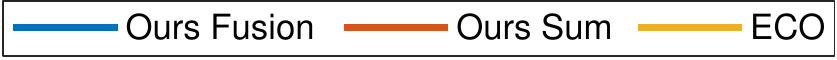}\vspace{-3mm}
	\caption{Analysis of tracking robustness and accuracy using the OP scores at IOU thresholds of $0.5$ and $0.75$ respectively on the NFS and OTB-E datasets. We plot the performance of our approach using sum-fusion with fixed weights (red) for a range of different shallow weights $\beta_\shallow$. These results are also compared with the baseline ECO (orange) and our adaptive fusion (blue). For a wide range of $\beta_\shallow$ values, our sum-fusion approach outperforms the baseline ECO in robustness on both datasets. Our adaptive fusion achieves the best performance both in terms of accuracy \emph{and} robustness.}\vspace{-4mm}
	\label{fig:naive_sum_baseline}
\end{figure}

We first investigate the impact of the observations from section~\ref{sec:analysis} in a tracking framework employing both deep and shallow features. To independently evaluate our contributions, we fuse the model predictions as in \eqref{eq:fusionsum} with fixed weights $\beta$. By varying these weights, we can further analyze the contribution of the deep and shallow models to the final tracking accuracy and robustness. We employ the widely used PASCAL criterion as an indicator of robustness. It measures the percentage of successfully tracked frames using an IOU threshold of $0.5$, equivalent to OP at $0.5$. Furthermore, we consider a localization to be \emph{accurate} if its IOU is higher than $0.75$, since this is the upper half $[0.75,1]$ of the IOU range $[0.5,1]$ representing successfully tracked frames.

Figure~\ref{fig:naive_sum_baseline} plots the accuracy and robustness indicators, as described above, on NFS and OTB-E for different values of the shallow model weight $\beta_\shallow$. In all cases, the deep weight is set to $\beta_\deep = 1-\beta_\shallow$. We also show the performance of the baseline ECO, using the \emph{same} set of deep and shallow features. We observe that our tracker with a fixed sum-fusion outperforms the baseline ECO for a wide range of weights $\beta_\shallow$. This demonstrates the importance of employing specifically tailored training procedures for deep and shallow features, as observed in section~\ref{sec:observations}.

\begin{figure}[t]
	\centering
	\begin{subfigure}[t]{0.2\textwidth}
		\includegraphics[width=\textwidth,trim={3cm 0cm 3cm 0cm},clip]{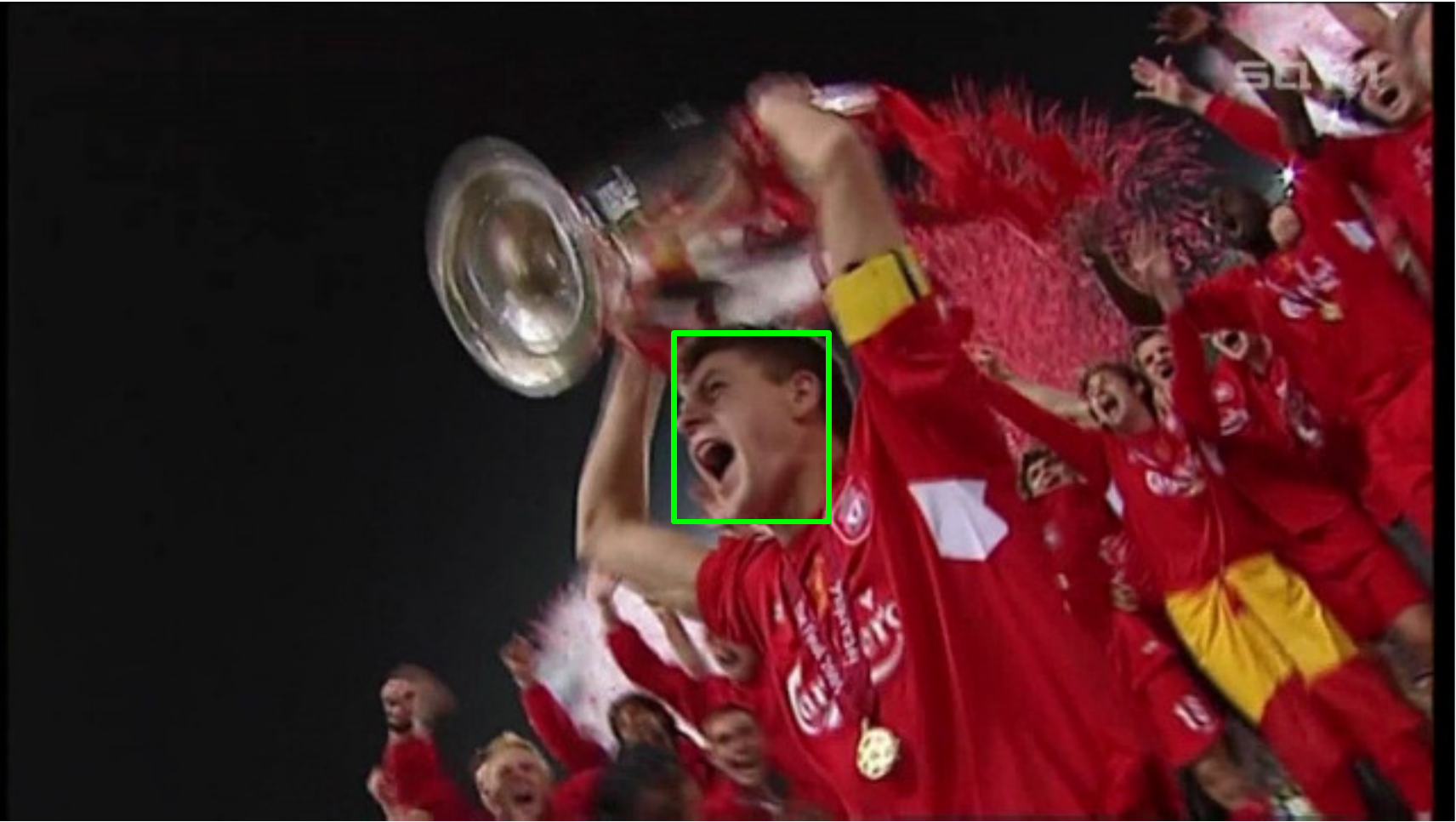}\vspace{-1.5mm}
		\subcaption{$\beta_d=0.01$}{\hspace{0.8cm}$\beta_s=0.99$}
	\end{subfigure}~%
	\begin{subfigure}[t]{0.2\textwidth}
		\includegraphics[width=\textwidth,trim={3cm 0cm 3cm 0cm},clip]{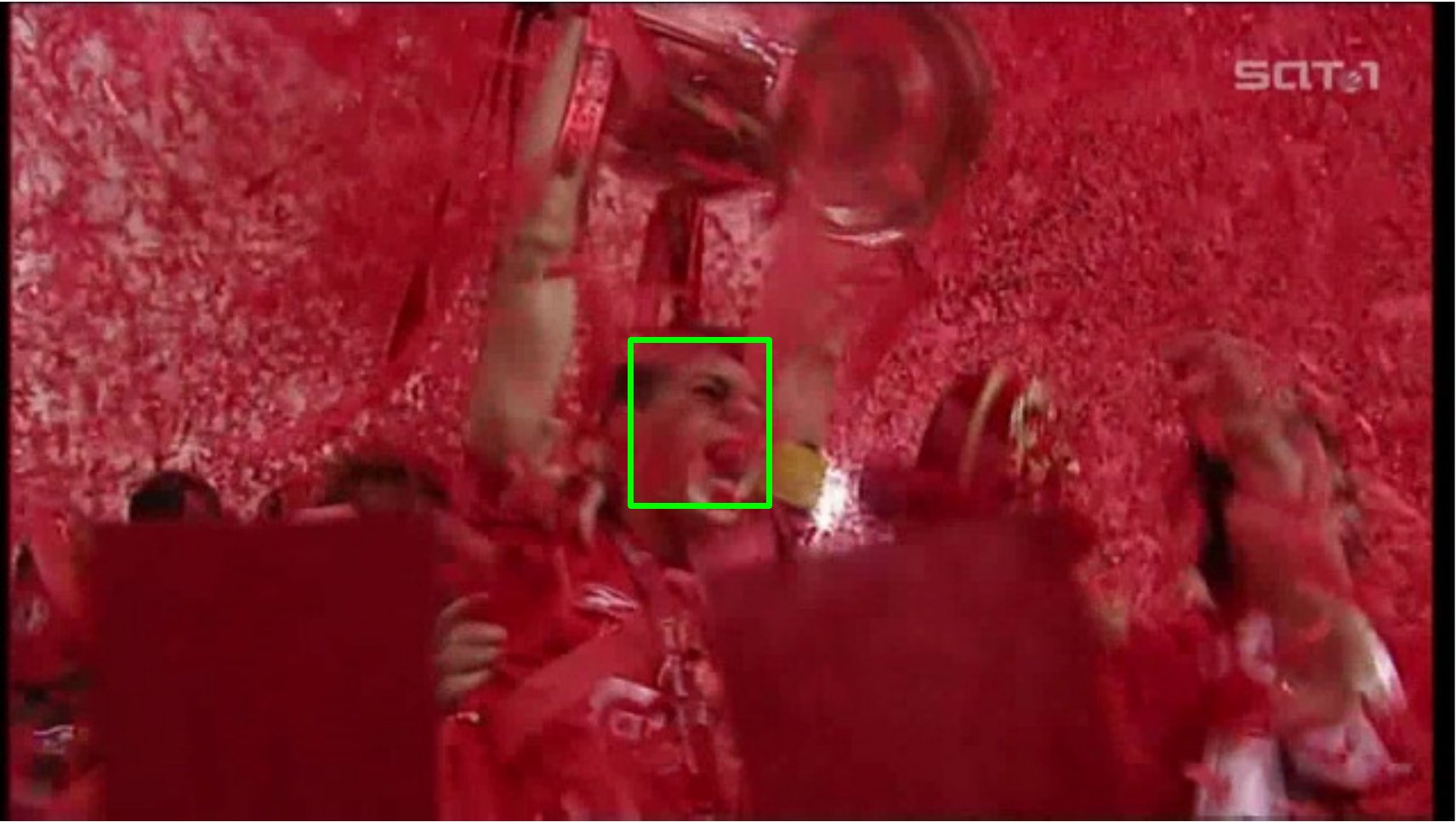}\vspace{-1.5mm}
		\subcaption{$\beta_d=0.90$}{\hspace{0.8cm}$\beta_s=0.10$}
	\end{subfigure}~%
	\begin{subfigure}[t]{0.2\textwidth}
		\includegraphics[width=\textwidth,trim={3cm 0cm 3cm 0cm},clip]{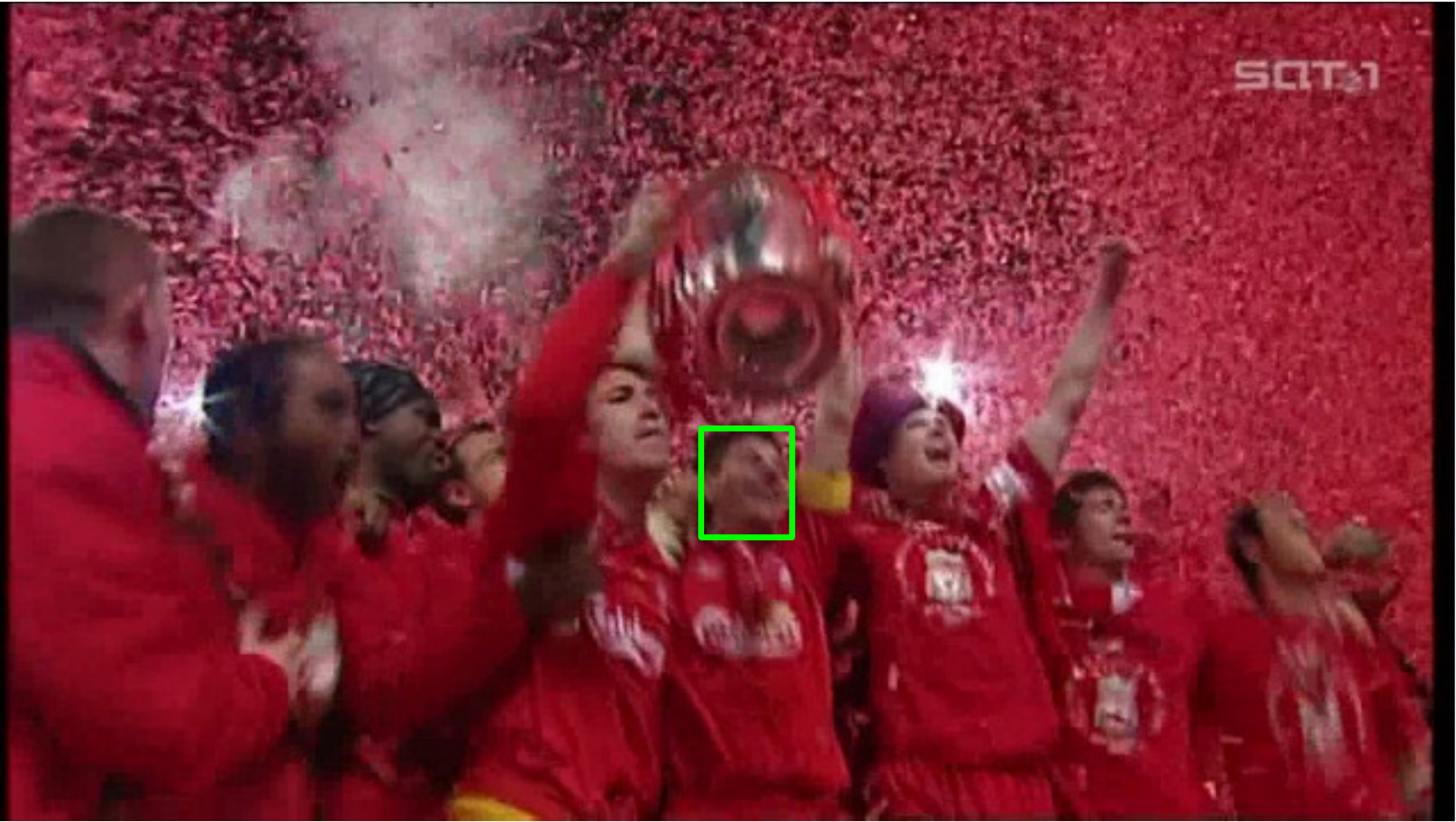}\vspace{-1.5mm}
		\subcaption{$\beta_d=0.10$}{\hspace{0.8cm}$\beta_s=0.90$}
	\end{subfigure}~%
	\begin{subfigure}[t]{0.2\textwidth}
		\includegraphics[width=\textwidth,trim={3cm 0cm 3cm 0cm},clip]{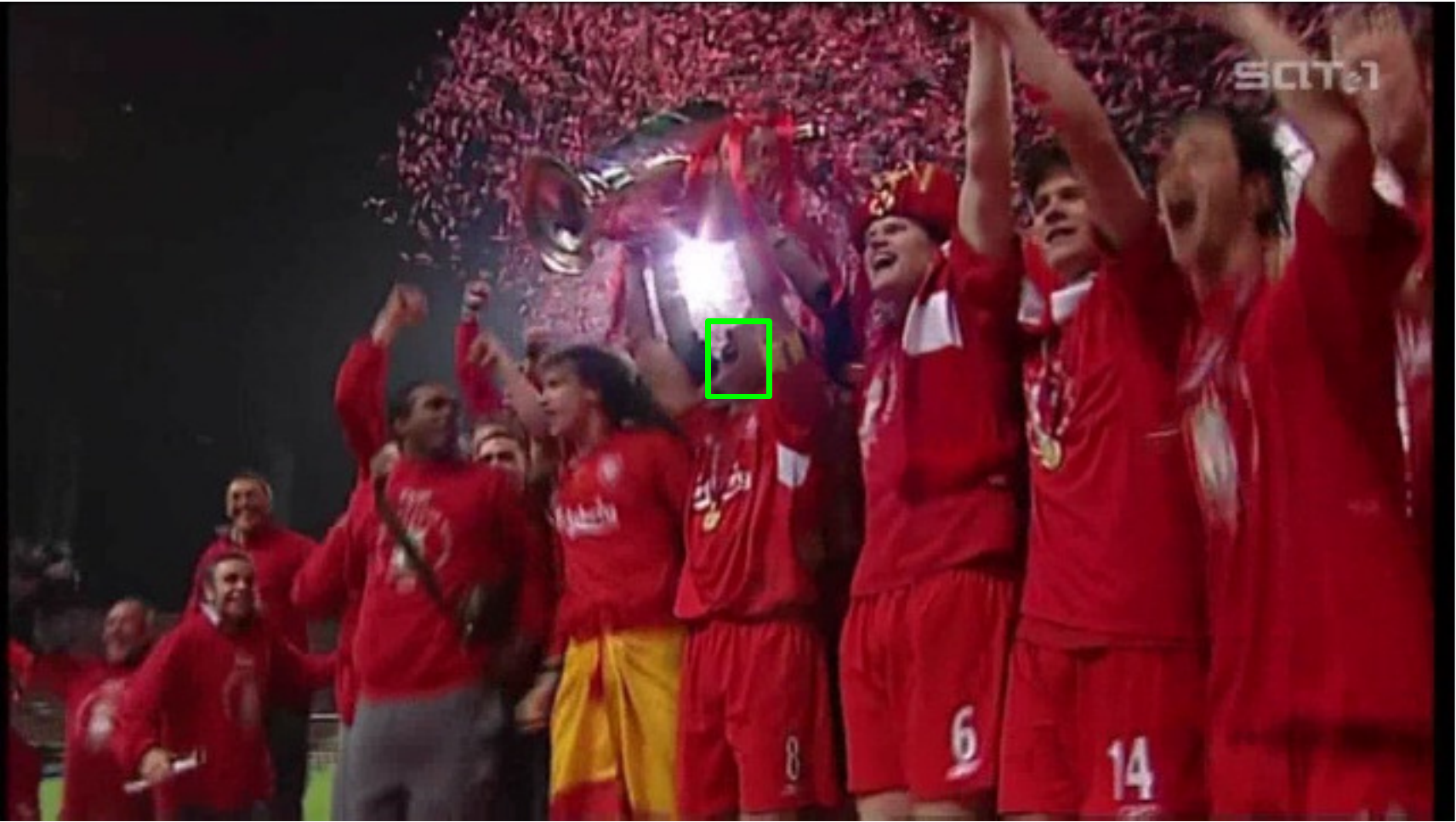}\vspace{-1.5mm}
		\subcaption{$\beta_d=0.87$}{\hspace{0.8cm}$\beta_s=0.13$}
	\end{subfigure}\vspace{-3mm}
	\caption{Qualitative example of our fusion approach. The adaptively computed model weights $\beta_d,\beta_s$ are shown for four frames from the \emph{Soccer} sequence. The shallow model is prominent early in the sequence (a), before any significant appearance change. Later, when encountered with occlusions, clutter and out-of-plane rotations (b,d), our fusion emphasizes the deep model due to its superior robustness. In (c), where the target undergoes scale changes, our fusion exploits the shallow model for better accuracy.}\vspace{-4mm}
	\label{fig:model_weights}
\end{figure}

Despite the above improvements obtained by our analysis of deep and shallow features, we note that optimal robustness and accuracy are mutually exclusive, and cannot be obtained even by careful selection of the weight parameter $\beta_\shallow$. While shallow features (large $\beta_\shallow$) are beneficial for accuracy, deep features (small $\beta_\shallow$) are crucial for robustness. Figure~\ref{fig:naive_sum_baseline} also shows the results of our proposed adaptive fusion approach (section~\ref{sec:fusion}), where the model weights $\beta$ are dynamically computed in each frame. Compared to using a sum-fusion with fixed weights, our adaptive approach achieves improved accuracy \emph{without} sacrificing robustness, using the same parameter settings for all datasets. Figure~\ref{fig:model_weights} shows a qualitative example of our adaptive fusion approach.

\begin{table}[!t]
	\centering%
	\caption{Generalization of our tracker across different network architectures. Results are shown in terms of AUC scores on the NFS and OTB-E dataset. The baseline ECO fails to exploit the power of more sophisticated architectures. Instead, our approach provides consistent gains over ECO when moving towards more advanced networks.}
	\resizebox{0.65\columnwidth}{!}{%
		\begin{tabular}{l @{~~~~~~} c @{~~} c @{~~~~~~} c @{~~} c @{~~~~~~} c @{~~} c}
\toprule
& \multicolumn{2}{c@{~~~~~~}}{\textbf{VGG-M}} & \multicolumn{2}{c@{~~~~~~~}}{\textbf{GoogLeNet}} & \multicolumn{2}{c@{~~}}{\textbf{ResNet-50}} \\
&OTB-E & NFS &OTB-E & NFS &OTB-E & NFS \\\midrule
ECO &74.8 & 45.3 &74.4 & 45.4 &74.3 & 45.7 \\
\textbf{Ours} &74.2 & 49.7 &76.0 & 51.6 &78.0 & 54.1 \\\bottomrule
\end{tabular}

	}
	\label{tab:network_generalization}%
	\vspace{-4mm}
\end{table}

\subsection{Generalization to Other Networks}
\label{sec:other_networks}
With the advent of deep learning, numerous network architectures have been proposed in recent years. Here, we investigate the generalization capabilities of our findings across different deep networks. Table~\ref{tab:network_generalization} shows the performance of the proposed method and baseline ECO on three popular architectures: VGG-M \cite{VGGM}, GoogLeNet \cite{Googlenet}, and ResNet-50 \cite{Resnet}. The results are reported in terms of AUC scores on both the NFS and OTB-E dataset.  
The baseline ECO tracker fails to exploit more sophisticated deeper architectures: GoogLeNet and ResNet. In case of ResNet architecture, our approach achieves a significant gain of $3.7\%$ and $8.4\%$ on OTB-E and NFS datasets respectively. These results demonstrate that our analysis in section~\ref{sec:analysis} and the fusion approach proposed in section~\ref{sec:fusion} generalizes across different network architectures.

\subsection{State-of-the-Art}
\label{sec:sota}

Here, we compare our tracker with state-of-the-art methods on four challenging tracking datasets. Detailed results are provided in the supplementary material.

\noindent\textbf{VOT2017 Dataset:} On VOT2017, containing 60 videos, tracking performance is evaluated both in terms of accuracy (average overlap during successful tracking) and robustness (failure rate). The Expected Average Overlap (EAO) measure, which merges both accuracy and robustness, is then used to obtain the overall ranking. The evaluation metrics are computed as an average over 15 runs (see \cite{VOT2017} for further details). The results in table~\ref{tab:VOT_sota} are presented in terms of expected average overlap (EAO), robustness, and accuracy. For clarity, we show the comparison with the top-10 best trackers in the VOT2017 competition.

The top performer LSART in the VOT2017 challenge, based on DCF and deep features, achieves an EAO score of $0.323$. 
Our approach significantly outperforms the best tracker in the VOT2017 challenge (LSART) with a relative gain of $17\%$, achieving an EAO score of $0.378$. In terms of robustness, our approach obtains a relative gain of $17\%$ compared to LSART. Furthermore, we achieve the best results in terms of accuracy, demonstrating its overall effectiveness.

\begin{figure}[t]
	\centering
	\begin{subfigure}[t]{0.32\textwidth}
		\includegraphics[width=1\textwidth,trim={0.0cm 0.0cm 0.0cm 0.0cm},clip]{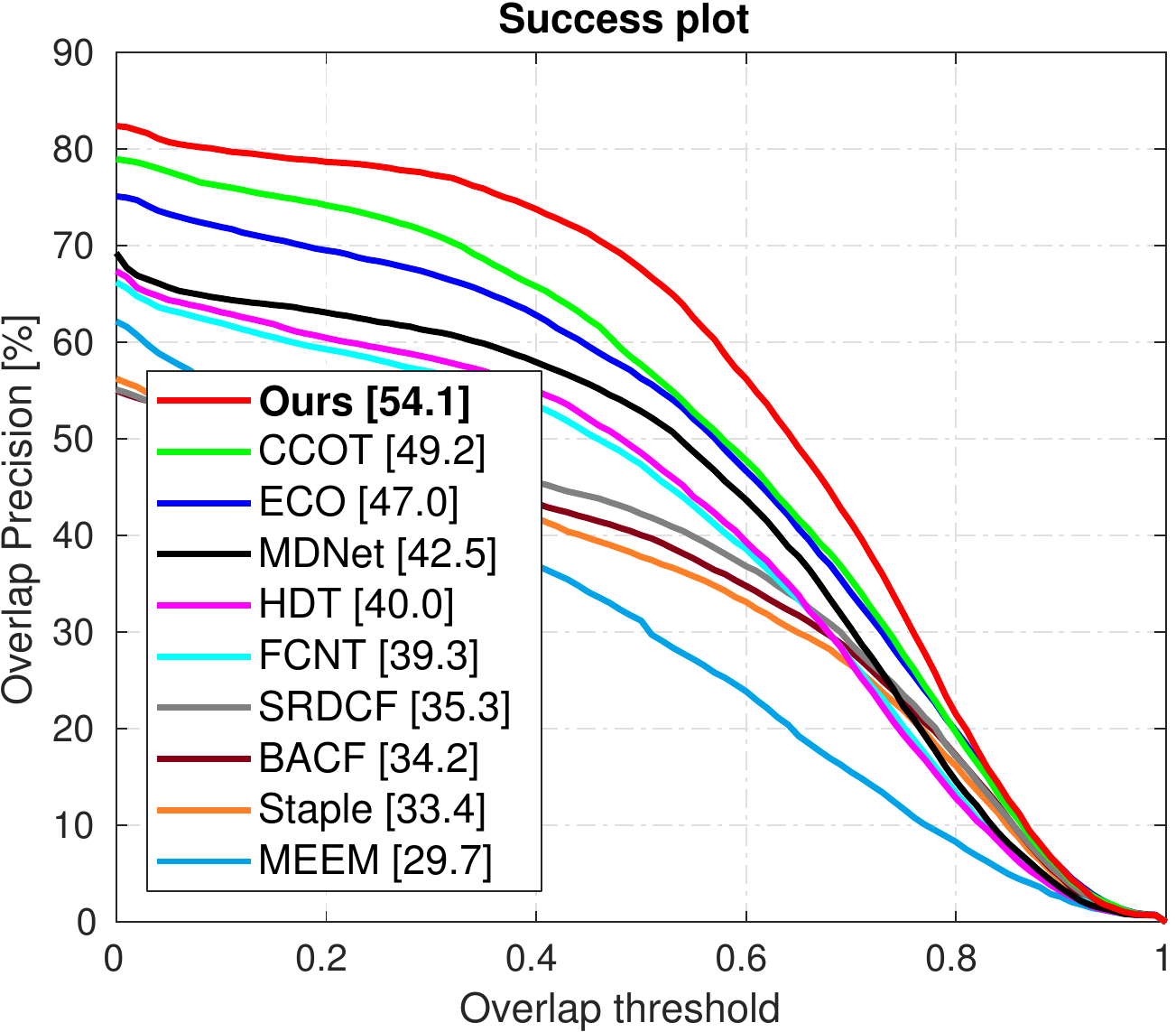}\vspace{-1mm}
		\caption{NFS}%
		\label{fig:nfs_success}
	\end{subfigure}\hspace{1mm}%
	\begin{subfigure}[t]{0.32\textwidth}
		\includegraphics[width=1\textwidth,trim={0.0cm 0.0cm 0.0cm 0.0cm},clip]{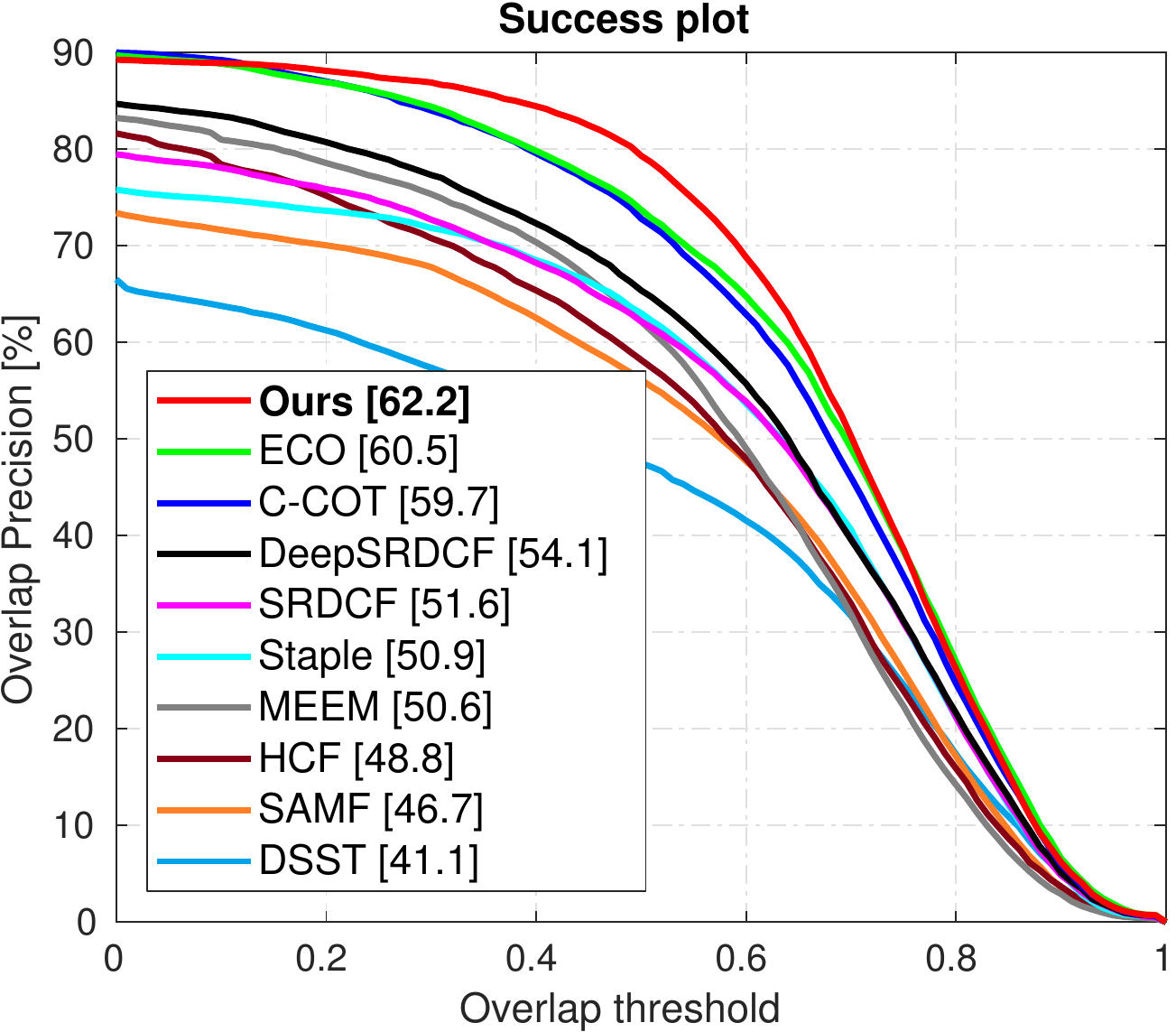}\vspace{-1mm}%
		\caption{Temple128}%
		\label{fig:sota_tpl}
	\end{subfigure}\hspace{1mm}%
	\begin{subfigure}[t]{0.32\textwidth}
		\includegraphics[width=1\textwidth,trim={0.0cm 0.0cm 0.0cm 0.0cm},clip]{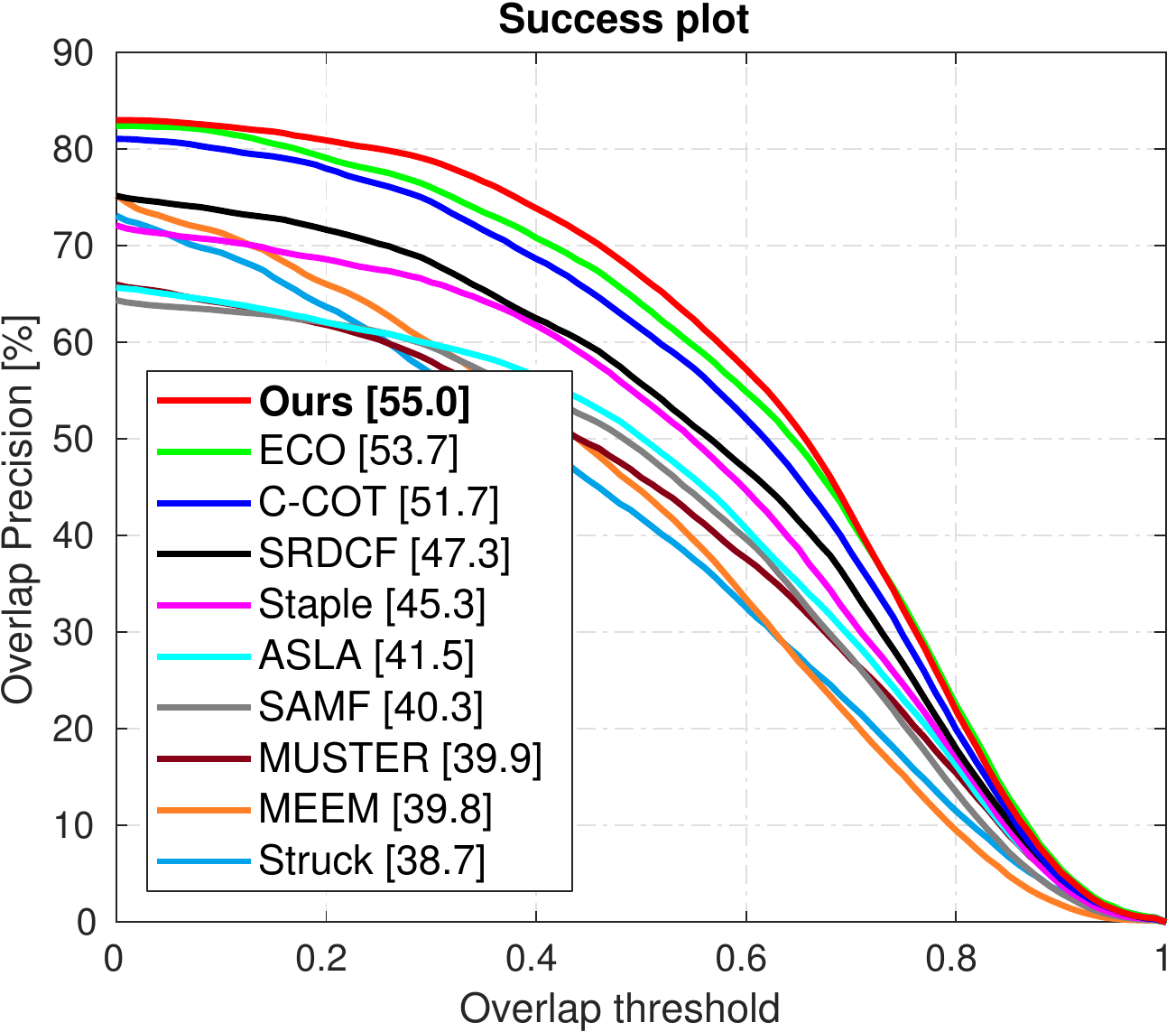}\vspace{-1mm}%
		\caption{UAV123}%
		\label{fig:sota_uav}
	\end{subfigure}%
	\vspace{-3mm}
	\caption{Success plots on the NFS (a), Temple128 (b), and UAV123 (c) datasets. For clarity, only the top 10 trackers are shown in the legend. Our tracker significantly outperforms the state-of-the-art on all datasets.}\vspace{-3mm}%
\end{figure}

\noindent\textbf{Need For Speed Dataset:} Figure~\ref{fig:nfs_success} shows the success plot over all the 100 videos. The AUC scores are reported in the legend. Among the existing methods, CCOT and ECO achieve AUC scores of $49.2\%$ and $47.0\%$ respectively. Our approach significantly outperforms CCOT with a relative gain of $10\%$.

\begin{table}[!t]
	\centering
	\caption{Comparison with the state-of-the-art in terms of expected average overlap (EAO), robustness (failure rate), and accuracy on the VOT2017 benchmark. We compare with the top-10 trackers in the competition. Our tracker obtains a significant relative gain of $17\%$ in EAO, compared to the top-ranked method (LSART).}%
	\label{tab:VOT_sota}%
	\resizebox{0.9\columnwidth}{!}{%
		\begin{tabular}{l@{~}c@{~~}c@{~~}c@{~~}c@{~~}c@{~~}c@{~~}c@{~~}c@{~~}c@{~~}c@{~~}c@{~~}}
\toprule
&MCPF&SiamDCF&CSRDCF&CCOT&MCCT&Gnet&ECO&CFCF&CFWCR&LSART&\textbf{Ours}\\\midrule
EAO&0.248&0.249&0.256&0.267&0.270&0.274&0.280&0.286&0.303&0.323&\tabfirst{0.378}\\
Robustness&0.427&0.473&0.356&0.318&0.323&0.276&0.276&0.281&0.267&0.218&\tabfirst{0.182}\\
Accuracy&0.510&0.500&0.491&0.494&0.525&0.502&0.483&0.509&0.484&0.493&\tabfirst{0.532}\\\bottomrule
\end{tabular}

	}\vspace{-1mm}%
\end{table}

\noindent\textbf{Temple128 Dataset:} Figure~\ref{fig:sota_tpl} shows the success plot over all 128 videos. Among the existing methods, ECO achieves an AUC score of $60.5\%$. Our approach outperforms ECO with an AUC score of $62.2\%$.

\noindent\textbf{UAV123 Dataset:} We evaluate our tracker on a dataset designed for low altitude UAV tracking, consisting of 123 videos. Figure~\ref{fig:sota_uav} shows the success plot. Among the existing methods, ECO achieves an AUC score of $53.7\%$. Our approach outperforms ECO by setting a new state-of-the-art, with an AUC of $55.0\%$.
\section{Conclusions}
In this paper, we perform a systematic analysis to identify key causes behind the below-expected performance of deep features for visual tracking. Our analysis shows that individually tailoring the training for shallow and deep features is crucial to obtain both high robustness and accuracy. We further propose a novel fusion strategy to combine the deep and shallow appearance models leveraging their complementary characteristics. Experiments are performed on four challenging datasets. Our experimental results clearly demonstrate the effectiveness of the proposed approach, leading to state-of-the-art performance on all datasets.

\bibliographystyle{splncs}
\bibliography{references}

\begin{thebibliography}{10}

\bibitem{DanelljanCVPR2017}
Danelljan, M., Bhat, G., Shahbaz~Khan, F., Felsberg, M.:
\newblock {ECO:} efficient convolution operators for tracking.
\newblock In: CVPR. (2017)

\bibitem{CSRDCF}
Luke{\v{z}}i{\v{c}}, A., Voj{\'\i}{\v{r}}, T., Zajc, L.{\v{C}}., Matas, J.,
  Kristan, M.:
\newblock Discriminative correlation filter tracker with channel and spatial
  reliability.
\newblock International Journal of Computer Vision  1--18

\bibitem{Staple}
Bertinetto, L., Valmadre, J., Golodetz, S., Miksik, O., Torr, P.H.S.:
\newblock Staple: Complementary learners for real-time tracking.
\newblock In: CVPR. (2016)

\bibitem{MEEM2014}
Zhang, J., Ma, S., Sclaroff, S.:
\newblock {MEEM:} robust tracking via multiple experts using entropy
  minimization.
\newblock In: ECCV. (2014)

\bibitem{OTB2015}
Wu, Y., Lim, J., Yang, M.H.:
\newblock Object tracking benchmark.
\newblock TPAMI \textbf{37}(9) (2015)  1834--1848

\bibitem{VOT2017}
Kristan, M., Leonardis, A., Matas, J., Felsberg, Pflugfelder, R., M.,
  \v{C}ehovin, L., Voj\'{i}r, T.and~Häger, G., et~al.\:
\newblock The visual object tracking vot2017 challenge results.
\newblock In: ICCV workshop. (2017)

\bibitem{DanelljanECCV2016}
Danelljan, M., Robinson, A., Khan, F., Felsberg, M.:
\newblock Beyond correlation filters: Learning continuous convolution operators
  for visual tracking.
\newblock In: ECCV. (2016)

\bibitem{HCF_ICCV15}
Ma, C., Huang, J.B., Yang, X., Yang, M.H.:
\newblock Hierarchical convolutional features for visual tracking.
\newblock In: ICCV. (2015)

\bibitem{NfS}
Galoogahi, H.K., Fagg, A., Huang, C., Ramanan, D., Lucey, S.:
\newblock Need for speed: A benchmark for higher frame rate object tracking.
\newblock In: 2017 IEEE International Conference on Computer Vision (ICCV),
  IEEE (2017)  1134--1143

\bibitem{Tao2016Sint}
Tao, R., Gavves, E., Smeulders, A.W.M.:
\newblock Siamese instance search for tracking.
\newblock In: CVPR. (2016)

\bibitem{LiBMVC14}
Li, H., Li, Y., Porikli, F.:
\newblock Deeptrack: Learning discriminative feature representations by
  convolutional neural networks for visual tracking.
\newblock In: BMVC. (2014)

\bibitem{Lijun15ICCV}
Wang, L., Ouyang, W., Wang, X., Lu, H.:
\newblock Visual tracking with fully convolutional networks.
\newblock In: ICCV. (2015)

\bibitem{MDNet}
Nam, H., Han, B.:
\newblock Learning multi-domain convolutional neural networks for visual
  tracking.
\newblock In: CVPR. (2016)

\bibitem{TCNN}
Nam, H., Baek, M., Han, B.:
\newblock Modeling and propagating cnns in a tree structure for visual
  tracking.
\newblock CoRR \textbf{abs/1608.07242} (2016)

\bibitem{Valmadre2017cvpr}
Valmadre, J., Bertinetto, L., Henriques, J.F., Vedaldi, A., Torr, P.H.S.:
\newblock End-to-end representation learning for correlation filter based
  tracking.
\newblock In: CVPR. (2017)

\bibitem{Song2017CREST}
Song, Y., Ma, C., Gong, L., Zhang, J., Lau, R., Yang, M.H.:
\newblock {CREST}: Convolutional residual learning for visual tracking.
\newblock In: ICCV. (2017)

\bibitem{MOSSE2010}
Bolme, D.S., Beveridge, J.R., Draper, B.A., Lui, Y.M.:
\newblock Visual object tracking using adaptive correlation filters.
\newblock In: CVPR. (2010)

\bibitem{DanelljanICCV2015}
Danelljan, M., H\"ager, G., Shahbaz~Khan, F., Felsberg, M.:
\newblock Learning spatially regularized correlation filters for visual
  tracking.
\newblock In: ICCV. (2015)

\bibitem{SiameseFC}
Bertinetto, L., Valmadre, J., Henriques, J.F., Vedaldi, A., Torr, P.H.:
\newblock Fully-convolutional siamese networks for object tracking.
\newblock In: ECCV workshop. (2016)

\bibitem{Qi2016Hedge}
Qi, Y., Zhang, S., Qin, L., Yao, H., Huang, Q., Lim, J., Yang, M.H.:
\newblock Hedged deep tracking.
\newblock In: CVPR. (2016)

\bibitem{ZhangCVPR17MCPF}
Zhang, T., Xu, C., Yang, M.H.:
\newblock Multi-task correlation particle filter for robust object tracking.
\newblock In: CVPR. (2017)

\bibitem{Dalal05}
Dalal, N., Triggs, B.:
\newblock Histograms of oriented gradients for human detection.
\newblock In: CVPR. (2005)

\bibitem{Weijer09a}
van~de Weijer, J., Schmid, C., Verbeek, J.J., Larlus, D.:
\newblock Learning color names for real-world applications.
\newblock TIP \textbf{18}(7) (2009)  1512--1524

\bibitem{li2014scale}
Li, Y., Zhu, J.:
\newblock A scale adaptive kernel correlation filter tracker with feature
  integration.
\newblock In: European Conference on Computer Vision, Springer (2014)  254--265

\bibitem{hong2015multi}
Hong, Z., Chen, Z., Wang, C., Mei, X., Prokhorov, D., Tao, D.:
\newblock Multi-store tracker (muster): A cognitive psychology inspired
  approach to object tracking.
\newblock In: Proceedings of the IEEE Conference on Computer Vision and Pattern
  Recognition. (2015)  749--758

\bibitem{lukevzivc2016discriminative}
Luke{\v{z}}i{\v{c}}, A., Voj{\'\i}{\v{r}}, T., {\v{C}}ehovin, L., Matas, J.,
  Kristan, M.:
\newblock Discriminative correlation filter with channel and spatial
  reliability.
\newblock arXiv preprint arXiv:1611.08461 (2016)

\bibitem{KristanPAMI2016}
Kristan, M., Matas, J., Leonardis, A., Voj{\'{\i}}r, T., Pflugfelder, R.P.,
  Fern{\'{a}}ndez, G., Nebehay, G., Porikli, F., Cehovin, L.:
\newblock A novel performance evaluation methodology for single-target
  trackers.
\newblock TPAMI \textbf{38}(11) (2016)  2137--2155

\bibitem{matconvnet}
Vedaldi, A., Lenc, K.:
\newblock Matconvnet -- convolutional neural networks for matlab.
\newblock CoRR \textbf{abs/1412.4564} (2014)

\bibitem{UAV123}
Mueller, M., Smith, N., Ghanem, B.:
\newblock A benchmark and simulator for uav tracking.
\newblock In: ECCV. (2016)

\bibitem{TempleColor}
Liang, P., Blasch, E., Ling, H.:
\newblock Encoding color information for visual tracking: Algorithms and
  benchmark.
\newblock TIP \textbf{24}(12) (2015)  5630--5644

\bibitem{VGGM}
Chatfield, K., Simonyan, K., Vedaldi, A., Zisserman, A.:
\newblock Return of the devil in the details: Delving deep into convolutional
  nets.
\newblock arXiv preprint arXiv:1405.3531 (2014)

\bibitem{Googlenet}
Szegedy, C., Liu, W., Jia, Y., Sermanet, P., Reed, S.E., Anguelov, D., Erhan,
  D., Vanhoucke, V., Rabinovich, A.:
\newblock Going deeper with convolutions.
\newblock In: CVPR. (2015)

\bibitem{Resnet}
He, K., Zhang, X., Ren, S., Sun, J.:
\newblock Deep residual learning for image recognition.
\newblock In: Proceedings of the IEEE conference on computer vision and pattern
  recognition. (2016)  770--778

\bibitem{horn1990matrix}
Horn, R.A., Johnson, C.R.:
\newblock Matrix analysis.
\newblock Cambridge university press (1990)

\bibitem{VOT2016}
Kristan, M., Leonardis, A., Matas, J., Felsberg, Pflugfelder, R., M.,
  \v{C}ehovin, L., Voj\'{i}r, T.and~Häger, G., et~al.\:
\newblock The visual object tracking vot2016 challenge results.
\newblock In: ECCV workshop. (2016)

\bibitem{VITALCVPR2018}
Song, Y., Ma, C., Wu, X., Gong, L., Bao, L., Zuo, W., Shen, C., Lau, R., Yang,
  M.H.:
\newblock {VITAL}: Visual tracking via adversarial learning.
\newblock In: CVPR. (2018)

\bibitem{SASiamCVPR2018}
He, A., Luo, C., Tian, X., Zeng, W.:
\newblock A twofold siamese network for real-time object tracking.
\newblock In: CVPR. (2018)

\bibitem{FlowTrackCVPR2018}
Zhu, Z., Wu, W., Zou, W., Yan, J.:
\newblock End-to-end flow correlation tracking with spatial-temporal attention.
\newblock In: CVPR. (2018)

\bibitem{LSARTCVPR2018}
Sun, C., Lu, H., Yang, M.H.:
\newblock Learning spatial-aware regressions for visual tracking.
\newblock In: CVPR. (2018)

\bibitem{DanelljanVOT2015}
Danelljan, M., H\"ager, G., Shahbaz~Khan, F., Felsberg, M.:
\newblock Convolutional features for correlation filter based visual tracking.
\newblock In: ICCV Workshop. (2015)

\bibitem{BACFgaloogahi}
Kiani~Galoogahi, H., Fagg, A., Lucey, S.:
\newblock Learning background-aware correlation filters for visual tracking.
\newblock In: ICCV. (2017)

\bibitem{SAMF}
Li, Y., Zhu, J.:
\newblock A scale adaptive kernel correlation filter tracker with feature
  integration.
\newblock In: ECCV workshop. (2014)

\bibitem{danelljan2016discriminative}
Danelljan, M., Hager, G., Khan, F.S., Felsberg, M.:
\newblock Discriminative scale space tracking.
\newblock IEEE Transactions on Pattern Analysis and Machine Intelligence (2016)

\bibitem{Torr11b}
Hare, S., Saffari, A., Torr, P.:
\newblock Struck: Structured output tracking with kernels.
\newblock In: ICCV. (2011)

\bibitem{Shengfeng13b}
He, S., Yang, Q., Lau, R., Wang, J., Yang, M.H.:
\newblock Visual tracking via locality sensitive histograms.
\newblock In: CVPR. (2013)

\end{thebibliography}

\clearpage
\setcounter{equation}{0}
\setcounter{figure}{0}
\setcounter{table}{0}
\setcounter{section}{0}

\renewcommand{\theequation}{S\arabic{equation}}
\renewcommand{\thefigure}{S\arabic{figure}}
\renewcommand{\thetable}{S\arabic{table}}
\renewcommand{\thesection}{S\arabic{section}}

\begin{center}
	\textbf{\large Supplementary Material}
\end{center}

In this supplementary material we provide additional derivation and results. Section \ref{sec:quality_measure} provides details about the derivation of eq. (4) in the paper. Section \ref{sec:otb_h_e} provides additional details about the OTB-H and OTB-E datasets introduced in section 5.2 of the main paper. Section \ref{sec:vot16_results} provides results on the VOT2016 dataset. Additional results on Need For Speed, UAV123, and Temple128 datasets are provided in section \ref{sec:additional_results}. Section \ref{sec:otb_results} provides results on the full OTB-2015 dataset.

\section{Derivation of Quality Measure Properties}
\label{sec:quality_measure}
In section 4.1 of the main paper, a quality measure was proposed as
\begin{align}
  &\xi_{t^*}\{y\} = \min_t\frac{y(t^*)-y(t)}{\Delta(t-t^*)},\text{ with }\\
  &\Delta(\tau) = 1-e^{-\frac{\kappa}{2}|\tau|^2}.
\end{align}
The quality is bounded from above,
\begin{align}
  \xi_{t^*}\{y\} \le \frac{|\lambda_1|}{\kappa},
\end{align}
where $\lambda_1$ is the largest eigenvalue of the Hessian of the detection score $\mathbf{H}y(t^*)$. In this section we show this.

First we assume that the detection score $y(t)$ is twice continuously differentiable. Let $\tau = t-t^*$, and note that the gradient at the local maximum $\nabla y(t^*)$ is zero. The corresponding Hessian $\mathbf{H}y(t^*)$ is negative semidefinite with eigenvalues $0\ge\lambda_1\ge\lambda_2$. Taylor expansions of the numerator and denominator results in
\begin{align}
  \xi_{t^*}\{y\} &= \min_\tau\frac{y(t^*)-\left(y(t^*)+\tau\tp\nabla y(t^*)+ \frac{1}{2}\tau\tp\mathbf{H}y(t^*)\tau + \ordo(|\tau|^3)\right)}{1-\left(1-\frac{\kappa}{2}|\tau|^2+\ordo(|\tau|^4)\right)}\\
  &= \min_\tau\frac{-\tau\tp\mathbf{H}y(t^*)\tau + \ordo(|\tau|^3)}{\kappa|\tau|^2+\ordo(|\tau|^4)}.
  \label{quality_taylor}
\end{align}
By considering the eigenvalue decomposition of the symmetric Hessian,
\begin{align}
  \lambda_2|\tau|^2\le\tau\tp\mathbf{H}\tau\le\lambda_1|\tau|^2,
\end{align}
where the bounds are obtained by the eigenvectors corresponding to $\lambda_2,\lambda_1$ \cite[\small\textit{4.2.2c}]{horn1990matrix}. For our case this means that if $\tau$ is an eigenvector with eigenvalue $\lambda_1$, we have that
\begin{align}
  -\tau\tp\mathbf{H}y(t^*)\tau=|\lambda_1||\tau|^2.
\end{align}
For such a $\tau$, the fraction in \eqref{quality_taylor} turns into
\begin{align}
  \frac{|\lambda_1||\tau|^2 + \ordo(|\tau|^3)}{\kappa|\tau|^2+\ordo(|\tau|^4)},
\end{align}
and its limit when $\tau$ approaches zero is
\begin{align}
  &\lim_{\tau\rightarrow 0} \frac{|\lambda_1||\tau|^2 + \ordo(|\tau|^3)}{\kappa|\tau|^2+\ordo(|\tau|^4)} = \frac{|\lambda_1|}{\kappa}.
\end{align}
As the quality definition aims at finding the minimum for all $\tau$, we obtain the bound
\begin{align}
  \xi_{t^*}\{y\} \le \frac{|\lambda_1|}{\kappa}.
  \label{eq:bound}
\end{align}
That is, the quality is bounded from above by the minimum curvature at the local optimum $t^*$. Note that the bound is tight, there are detection scores for which it is equal to the quality. Consider for instance the case where the detection score is a Gaussian centered at $p$
\begin{align}
  y(t) = e^{-\frac{\kappa}{4}|t-p|^2},
\end{align}
and where $\kappa=4$. The Hessian at the optimum $p$ is
\begin{align}
  \mathbf{H} y(t) = \begin{pmatrix}-\frac{\kappa}{2} & 0\\ 0 & -\frac{\kappa}{2}\end{pmatrix},
\end{align}
with eigenvalues $\lambda_1=\lambda_2=-\frac{\kappa}{2}$. The corresponding bound \eqref{eq:bound} is $\xi_{p}\{y\}\le \frac{1}{2}$. Letting $\tau=t-p$, the score quality is
\begin{align}
  \xi_{p}\{y\} &= \min_\tau\frac{1 - e^{-\frac{\kappa}{4}|\tau|^2}}{1 - e^{-2\frac{\kappa}{4}|\tau|^2}} =\\
  &= \min_\tau\frac{1 - e^{-\frac{\kappa}{4}|\tau|^2}}{(1 - e^{-\frac{\kappa}{4}|\tau|^2})(1+e^{-\frac{\kappa}{4}|\tau|^2})}=\\
  &= \min_\tau\frac{1}{1+e^{-\frac{\kappa}{4}|\tau|^2}}=\\
  &= \frac{1}{2}.
\end{align}
That is, there is always a detection score for which the bound is attained and it is therefore tight.

\section{Description of OTB-H and OTB-E}
\label{sec:otb_h_e}
In this section, we describe the OTB-H and OTB-E datasets (introduced in section 5.2 of the main paper). Both OTB-H and OTB-E are constructed from the OTB-2015 benchmark~\cite{OTB2015} using the per video results of four state-of-the-art trackers (see the paper for details). OTB-H consists of 23 videos and is used as an explicit validation set to set the hyperparameters. OTB-E consists of 73 videos and is used as test set for the ablation study.  Note that the sequences which are \textit{hard}, but overlap with the VOT2017 (Bolt2, Soccer, Matrix, MotorRolling) are excluded from both OTB-H and OTB-E. Following is the list of videos included in OTB-H and OTB-E. \\

\noindent\textbf{OTB-H:} Biker, Bird1, ClifBar, Coke, Coupon, Diving, Dog, Football1, Freeman1, Freeman4, Gym, Human3, Ironman, Jump, Panda, RedTeam, Skater, Skating1, Skating2\_1, Skating2\_2, Skiing, Trans, Vase \\

\noindent\textbf{OTB-E:} Basketball, Bird2, BlurBody, BlurCar1, BlurCar2, BlurCar3, BlurCar4, BlurFace, BlurOwl, Board, Bolt, Box, Boy, Car1, Car2, Car24, Car4, CarDark, CarScale, Couple, Crossing, Crowds, Dancer, Dancer2, David, David2, David3, Deer, Dog1, Doll, DragonBaby, Dudek, FaceOcc1, FaceOcc2, Fish, FleetFace, Football, Freeman3, Girl, Girl2, Human2, Human4\_2, Human5, Human6, Human7, Human8, Human9, Jogging\_1, Jogging\_2, Jumping, KiteSurf, Lemming, Liquor, Man, Mhyang, MountainBike, Rubik, Shaking, Singer1, Singer2, Skater2, Subway, Surfer, Suv, Sylvester, Tiger1, Tiger2, Toy, Trellis, Twinnings, Walking, Walking2, Woman \\

\begin{table}[!t]
	\centering
	\caption{Comparison with the state-of-the-art in terms of expected average overlap (EAO) on the VOT2016 dataset. Our tracker obtains an EAO score of $0.443$, outperforming the second best method (ECO) with relative gain of $18\%$.}%
	\label{tab:vot16}%
	\resizebox{0.9\columnwidth}{!}{%
		\begin{tabular}{l@{~}c@{~~}c@{~~}c@{~~}c@{~~}c@{~~}c@{~~}c@{~~}c@{~~}c@{~~}c@{~~}c@{~~}}
\toprule
&CREST&SA-Siam&Staple&VITAL&LSART&TCNN&CCOT&FlowTrack&ECO&\textbf{Ours}\\\midrule
EAO&0.283&0.291&0.295&0.323&0.324&0.325&0.331&0.334&0.374&\tabfirst{0.443}\\\bottomrule
\end{tabular}

	}\vspace{-1mm}%
\end{table}

\section{Results on VOT2016}
\label{sec:vot16_results}
In this section, we evaluate our method on the VOT2016 dataset~\cite{VOT2016}. We compare our method with the following state-of-the-art methods: CCOT~\cite{DanelljanECCV2016}, ECO~\cite{DanelljanCVPR2017}, TCNN~\cite{TCNN}, VITAL~\cite{VITALCVPR2018}, SA-Siam~\cite{SASiamCVPR2018}, FlowTrack~\cite{FlowTrackCVPR2018}, LSART~\cite{LSARTCVPR2018}, CREST~\cite{Song2017CREST}, and Staple~\cite{Staple}. Results are shown in Table \ref{tab:vot16} in terms of Expected Average Overlap (EAO) metric. Our method achieves an EAO score of $0.443$, significantly outperforming ECO with a relative gain of $18\%$.

\begin{figure}[t]
	\centering
	\begin{subfigure}[t]{0.32\textwidth}
		\includegraphics[width=1\textwidth,trim={0.0cm 0.0cm 0.0cm 0.0cm},clip]{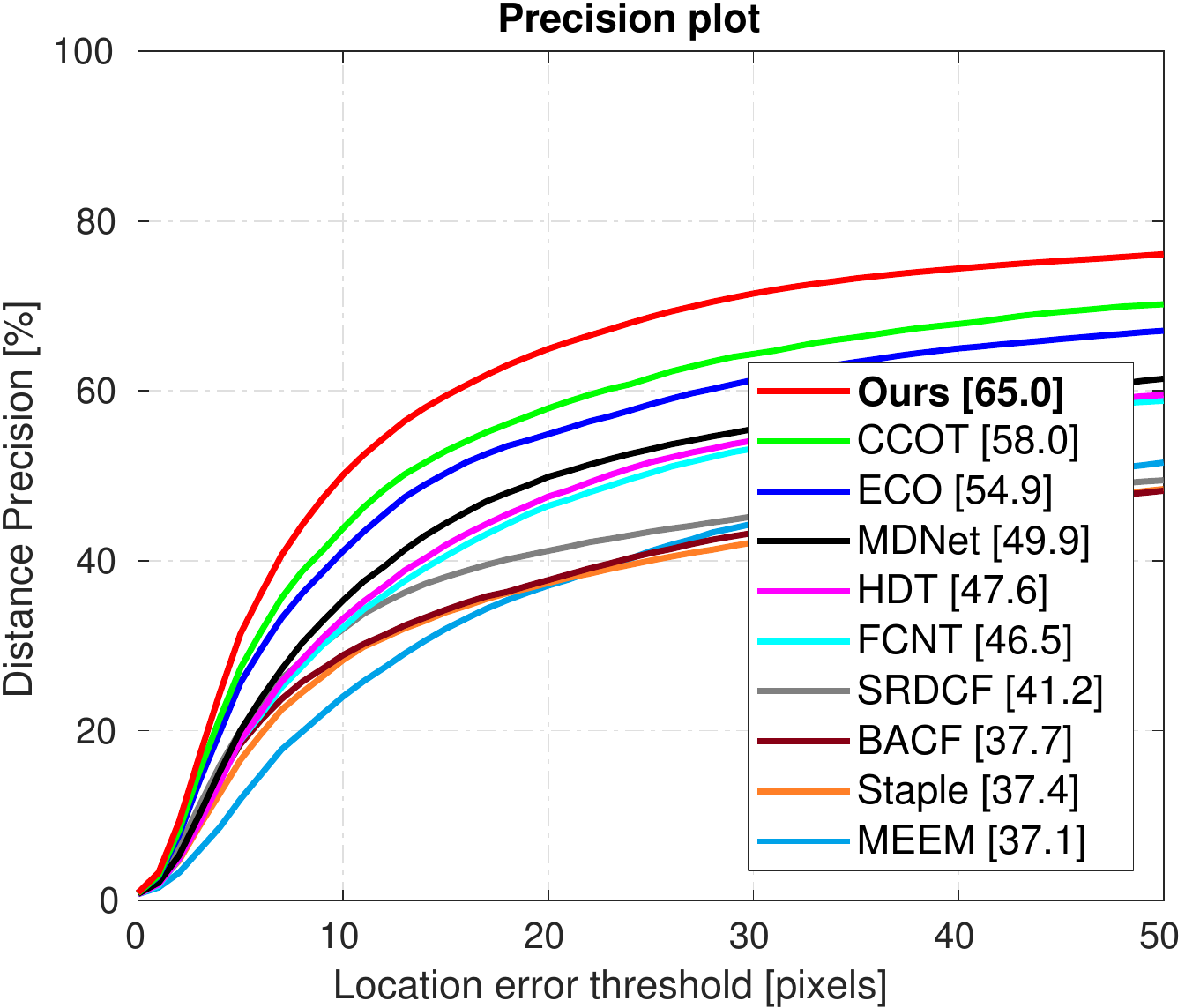}\vspace{-1mm}
		\caption{NFS}%
		\label{fig:nfs_prec}
	\end{subfigure}\hspace{1mm}%
	\begin{subfigure}[t]{0.32\textwidth}
		\includegraphics[width=1\textwidth,trim={0.0cm 0.0cm 0.0cm 0.0cm},clip]{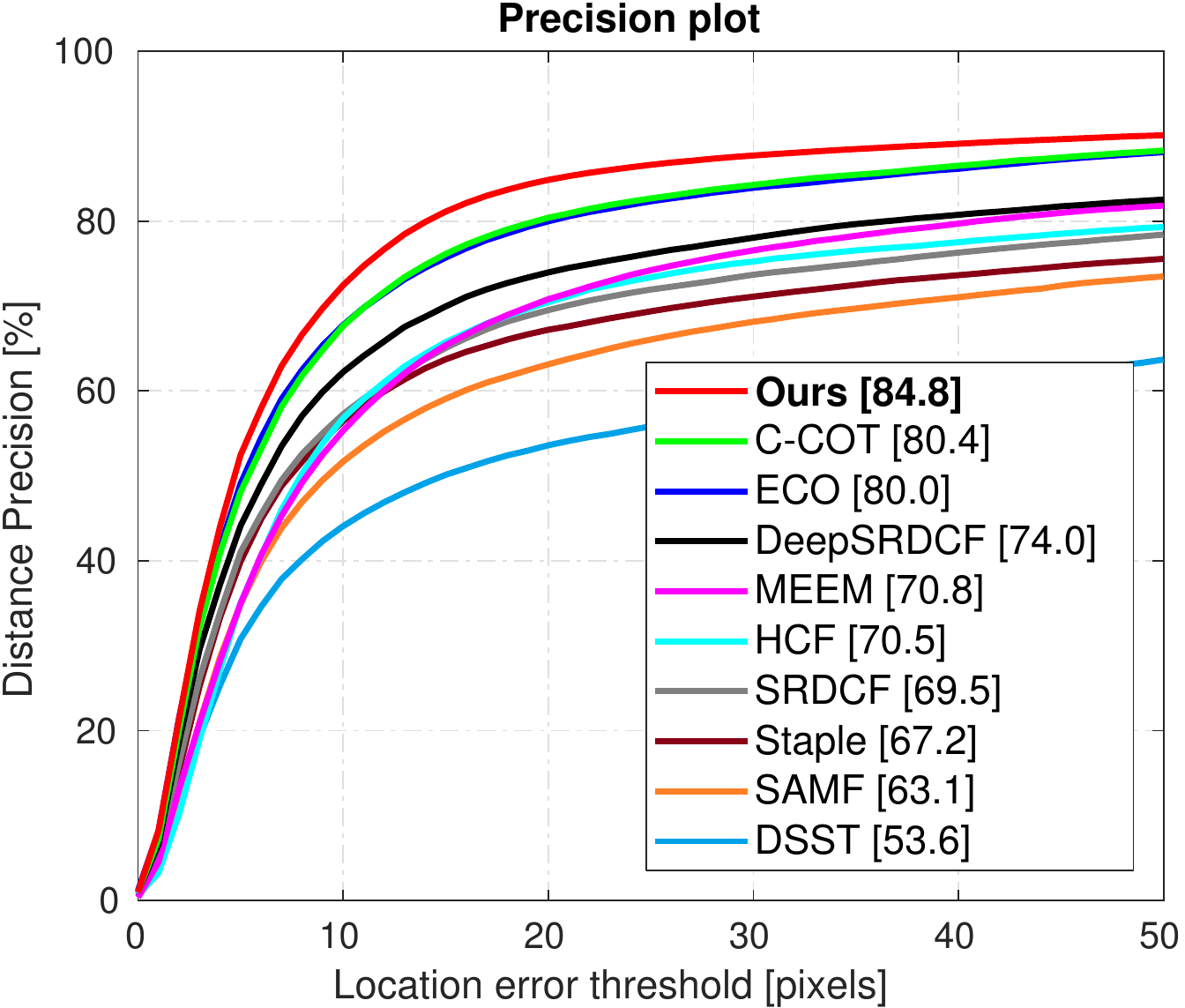}\vspace{-1mm}%
		\caption{Temple128}%
		\label{fig:tpl_prec}
	\end{subfigure}\hspace{1mm}%
	\begin{subfigure}[t]{0.32\textwidth}
		\includegraphics[width=1\textwidth,trim={0.0cm 0.0cm 0.0cm 0.0cm},clip]{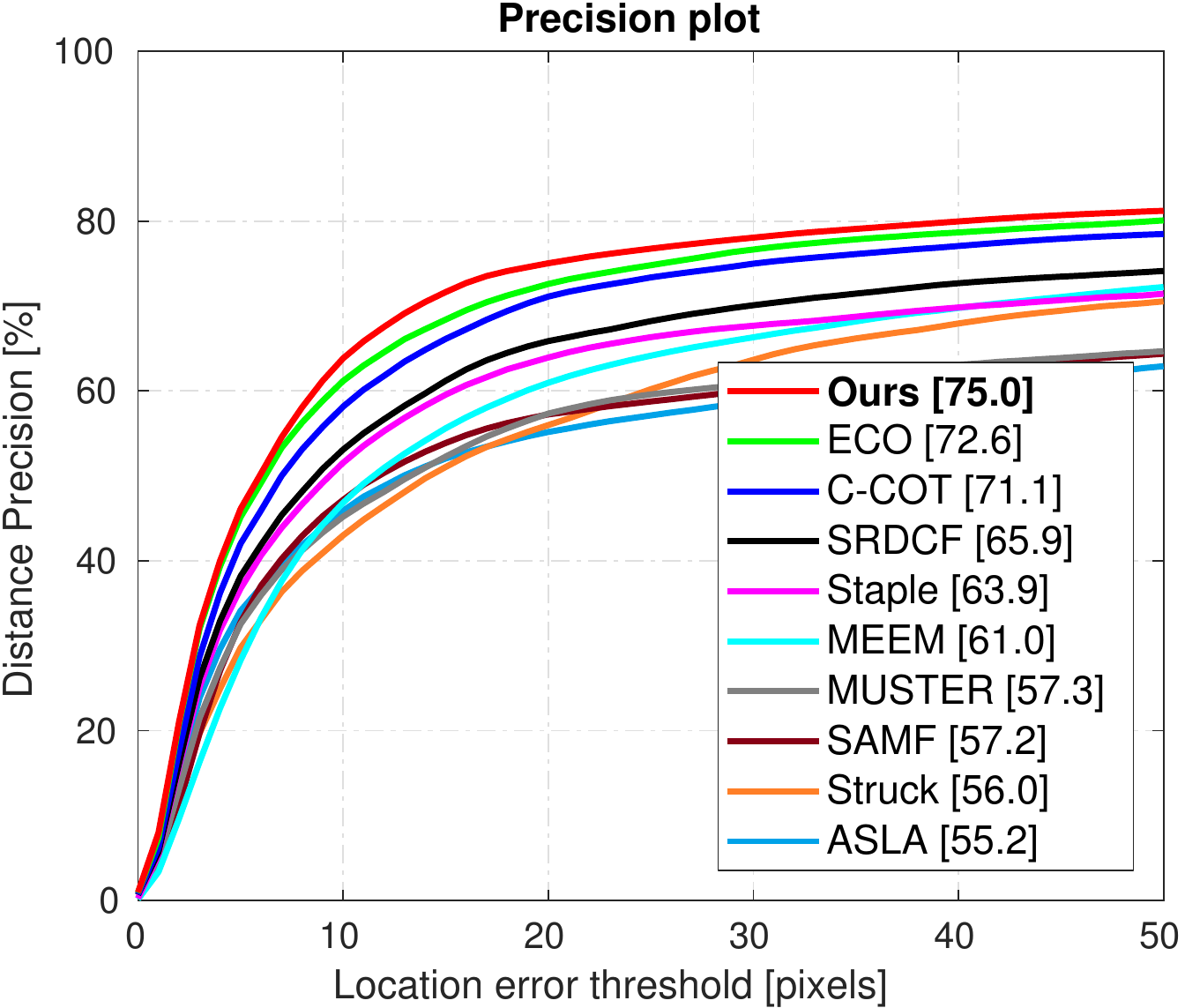}\vspace{-1mm}%
		\caption{UAV123}%
		\label{fig:uav_prec}
	\end{subfigure}%
	\vspace{-3mm}
	\caption{Precision plots on the NFS (a), Temple128 (b), and UAV123 (c) datasets. The DP scores for the top 10 trackers are shown in the legend. Our tracker significantly outperforms the state-of-the-art on all datasets.}
	\label{fig:sota_prec}
	\vspace{-3mm}%
\end{figure}

\section{Additional Results on NFS, UAV123 and Temple128}
\label{sec:additional_results}
In this section, we provide precision plots on Need For Speed (NFS)~\cite{NfS}, UAV123~\cite{UAV123}, and Temple128~\cite{TempleColor} datasets. We compare our method with the following state-of-the-art methods: CCOT~\cite{DanelljanECCV2016}, ECO~\cite{DanelljanCVPR2017}, MDNet~\cite{MDNet}, HDT~\cite{Qi2016Hedge}, FCNT~\cite{Lijun15ICCV}, SRDCF~\cite{DanelljanICCV2015}, DeepSRDCF~\cite{DanelljanVOT2015}, BACF~\cite{BACFgaloogahi}, Staple~\cite{Staple}, MEEM~\cite{MEEM2014}, HCF~\cite{HCF_ICCV15}, SAMF~\cite{SAMF}, DSST~\cite{danelljan2016discriminative}, MUSTER~\cite{hong2015multi}, Struck~\cite{Torr11b}, and ASLA~\cite{Shengfeng13b}. We use the distance precision (DP) scores to rank the trackers. DP is defined as the percentage of frames in which the Euclidean distance between the tracker prediction and the ground truth centers is less than a threshold (20). The DP score is plotted over a range of thresholds [0,50] to get the precision plot.  Figure \ref{fig:sota_prec} shows the precision plots on the three datasets. Our tracker significantly outperforms the state-of-the-art, achieving an absolute gain of $7.0\%$, $4.4\%$, and $2.4\%$ on NFS, Temple128, and UAV123 respectively.

\section{Results on OTB-2015}
\label{sec:otb_results}
\begin{figure}[t]
	\centering
	\includegraphics[width=0.5\textwidth,trim={0.0cm 0.0cm 0.0cm 0.0cm},clip]{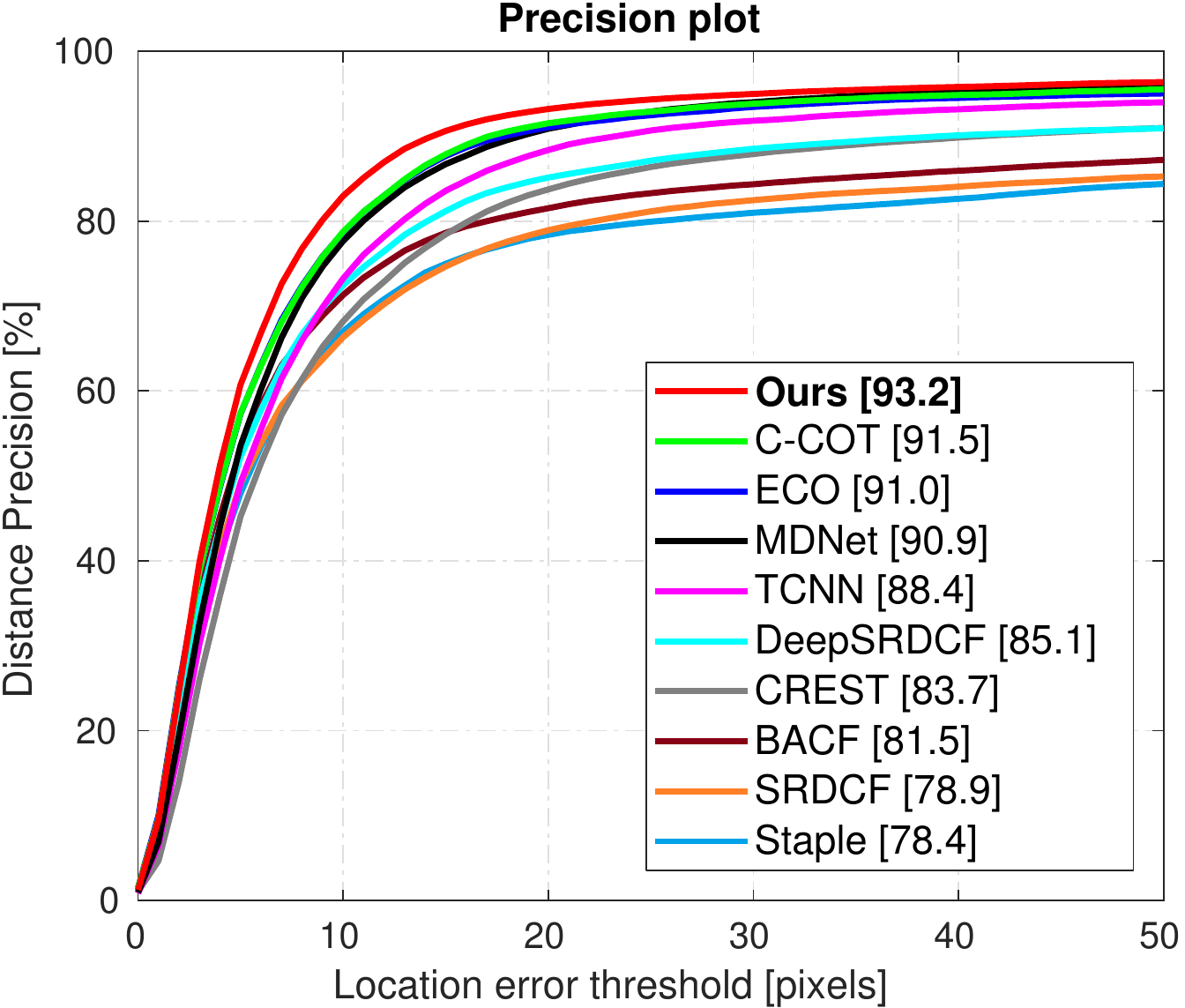}\vspace{-1mm}%
	\includegraphics[width=0.5\textwidth,trim={0.0cm 0.0cm 0.0cm 0.0cm},clip]{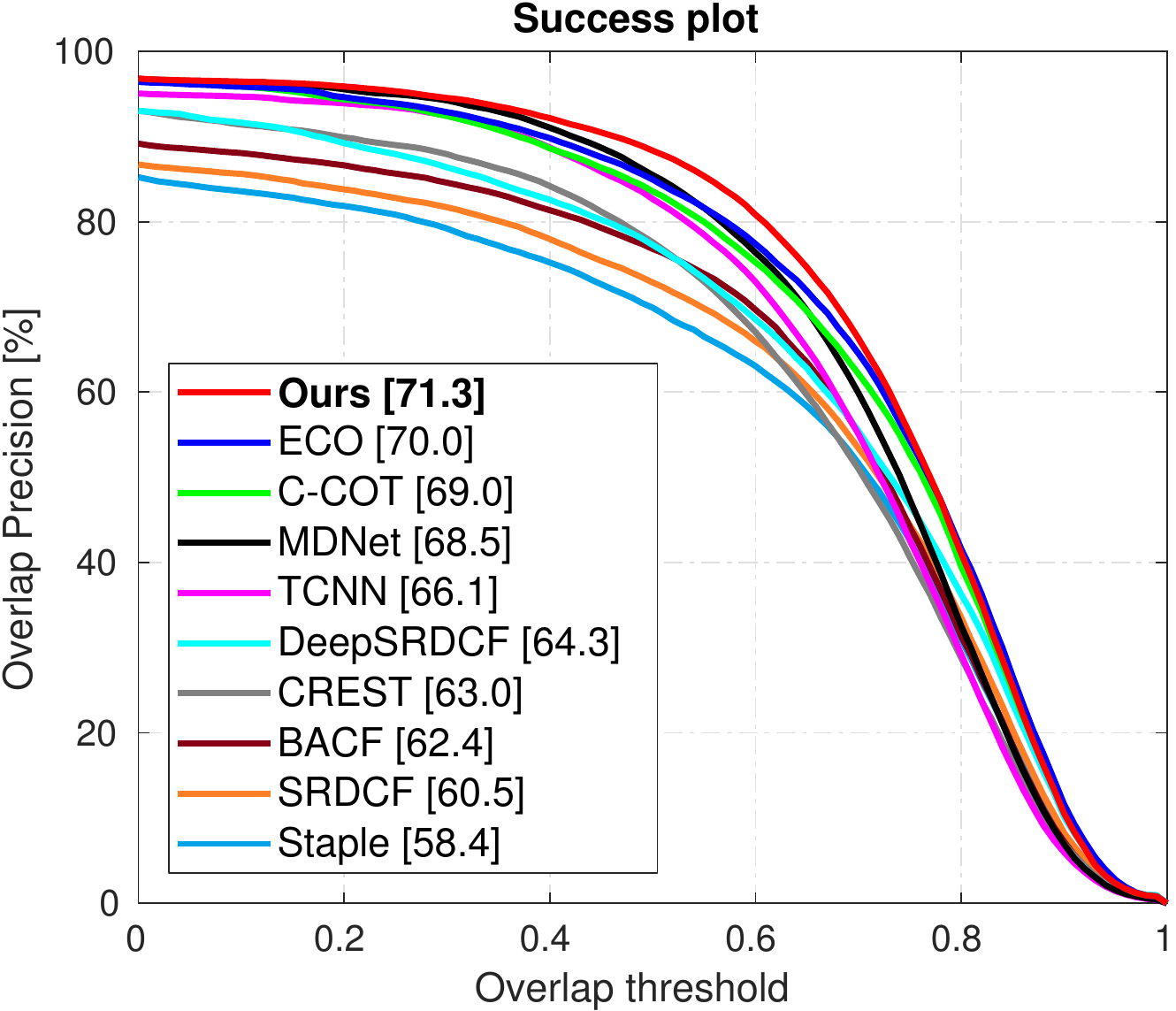}\vspace{-1mm}
	\vspace{-3mm}
	\caption{Precision (Left) and Success (Right) plots on the OTB-2015 dataset ~\cite{OTB2015}. The distance precision (DP) score and the area-under-the-curve (AUC) score are shown in the legend of Precison and Success plot, respectively. Our tracker achieves the best scores, both in terms of DP and AUC}\vspace{-3mm}
	\label{fig:otb}
\end{figure}

The performance of the proposed method on a subset of 73 videos from OTB-2015~\cite{OTB2015} (OTB-E) was provided in section 5.3 and section 5.4 of the main paper. For completeness, we also report the results on the full OTB-2015 dataset. Note that a subset of 23 videos from OTB-2015 (OTB-H) was used to set the hyperparameters of the proposed method. We compare our method with the following state-of-the-art methods: CCOT~\cite{DanelljanECCV2016}, ECO~\cite{DanelljanCVPR2017}, MDNet~\cite{MDNet}, TCNN~\cite{TCNN}, DeepSRDCF~\cite{DanelljanVOT2015}, CREST~\cite{Song2017CREST}, BACF~\cite{BACFgaloogahi}, SRDCF~\cite{DanelljanICCV2015}, and Staple~\cite{Staple}. We use the area-under-the-curve (AUC) and distance precision (DP) scores to rank the trackers. 
Figure \ref{fig:otb} contains the precision and success plots over all the 100 videos. Our method achieves the best scores in terms of both AUC and DP measures, achieving an absolute gain of $1.3\%$ and $1.7\%$ respectively.


\end{document}